\documentclass[10pt,twocolumn,letterpaper]{article}

\usepackage{iccv}
\usepackage{times}
\usepackage{epsfig}
\usepackage{graphicx}
\usepackage{amsmath}
\usepackage{amssymb}
\usepackage{multirow}
\usepackage{caption}
\usepackage{booktabs}
\usepackage{tabulary}
\usepackage{makecell}
\usepackage{colortbl}
\usepackage{color}
\usepackage{enumitem}
\usepackage{bbding}
\usepackage[accsupp]{axessibility}
\definecolor{mygray}{gray}{.93}
\usepackage{pifont}

\usepackage[breaklinks=true,colorlinks,bookmarks=false]{hyperref}

\iccvfinalcopy 

\def\iccvPaperID{4954} 

\ificcvfinal\fi

\begin{document}

\title{Scale-Aware Modulation Meet Transformer}
\author{\textbf{Weifeng Lin}\textsuperscript{\rm 1,2},\; \textbf{Ziheng Wu}\textsuperscript{\rm 2}, \textbf{Jiayu Chen}\textsuperscript{\rm 2}, \textbf{Jun Huang} \textsuperscript{\rm 2}, \textbf{Lianwen Jin}\textsuperscript{\rm 1}\thanks{Corresponding authors.} \vspace{0.2cm} \\
    \textsuperscript{\rm 1} South China University of Technology, 
    \textsuperscript{\rm 2} Platform of AI (PAI), Alibaba Group \\
\centerline{  \small
eelinweifeng@mail.scut.edu.cn
\quad eelwjin@scut.edu.cn
\quad \{ziheng.wzh, yunji.cjy, huangjun.hj\}@alibaba-inc.com } \vspace{-0.1cm}  \\
}

\maketitle
\ificcvfinal\thispagestyle{empty}\fi

\begin{abstract} 
   This paper presents a new vision Transformer, Scale-Aware Modulation Transformer (SMT), that can handle various downstream tasks efficiently by combining the convolutional network and vision Transformer. The proposed Scale-Aware Modulation (SAM) in the SMT includes two primary novel designs. Firstly, we introduce the Multi-Head Mixed Convolution (MHMC) module, which can capture multi-scale features and expand the receptive field.
   Secondly, we propose the Scale-Aware Aggregation (SAA) module, which is lightweight but effective, enabling information fusion across different heads. By leveraging these two modules, convolutional modulation is further enhanced. 
   Furthermore, in contrast to prior works that utilized modulations throughout all stages to build an attention-free network, we propose an Evolutionary Hybrid Network (EHN), which can effectively simulate the shift from capturing local to global dependencies as the network becomes deeper, resulting in superior performance. Extensive experiments demonstrate that SMT significantly outperforms existing state-of-the-art models across a wide range of visual tasks. 
   Specifically, SMT with \textbf{11.5M / 2.4GFLOPs} and \textbf{32M / 7.7GFLOPs} can achieve \textbf{82.2\%} and \textbf{84.3\%} top-1 accuracy on ImageNet-1K, respectively. After pretrained on ImageNet-22K in 224${^2}$ resolution, it attains \textbf{87.1\%} and \textbf{88.1\%} top-1 accuracy when finetuned with resolution 224${^2}$ and 384${^2}$, respectively.
   For object detection with Mask R-CNN, the SMT base trained with 1$\times$ and 3$\times$ schedule outperforms the Swin Transformer counterpart by \textbf{4.2} and \textbf{1.3} mAP on COCO, respectively. For semantic segmentation with UPerNet, the SMT base test at single- and multi-scale surpasses Swin by \textbf{2.0} and \textbf{1.1} mIoU respectively on the ADE20K. Our code is available at \url{https://github.com/AFeng-x/SMT}.
   
\end{abstract}

\begin{figure}[t]
    \centering
    \hspace{-1mm}\includegraphics[width=0.48\textwidth]{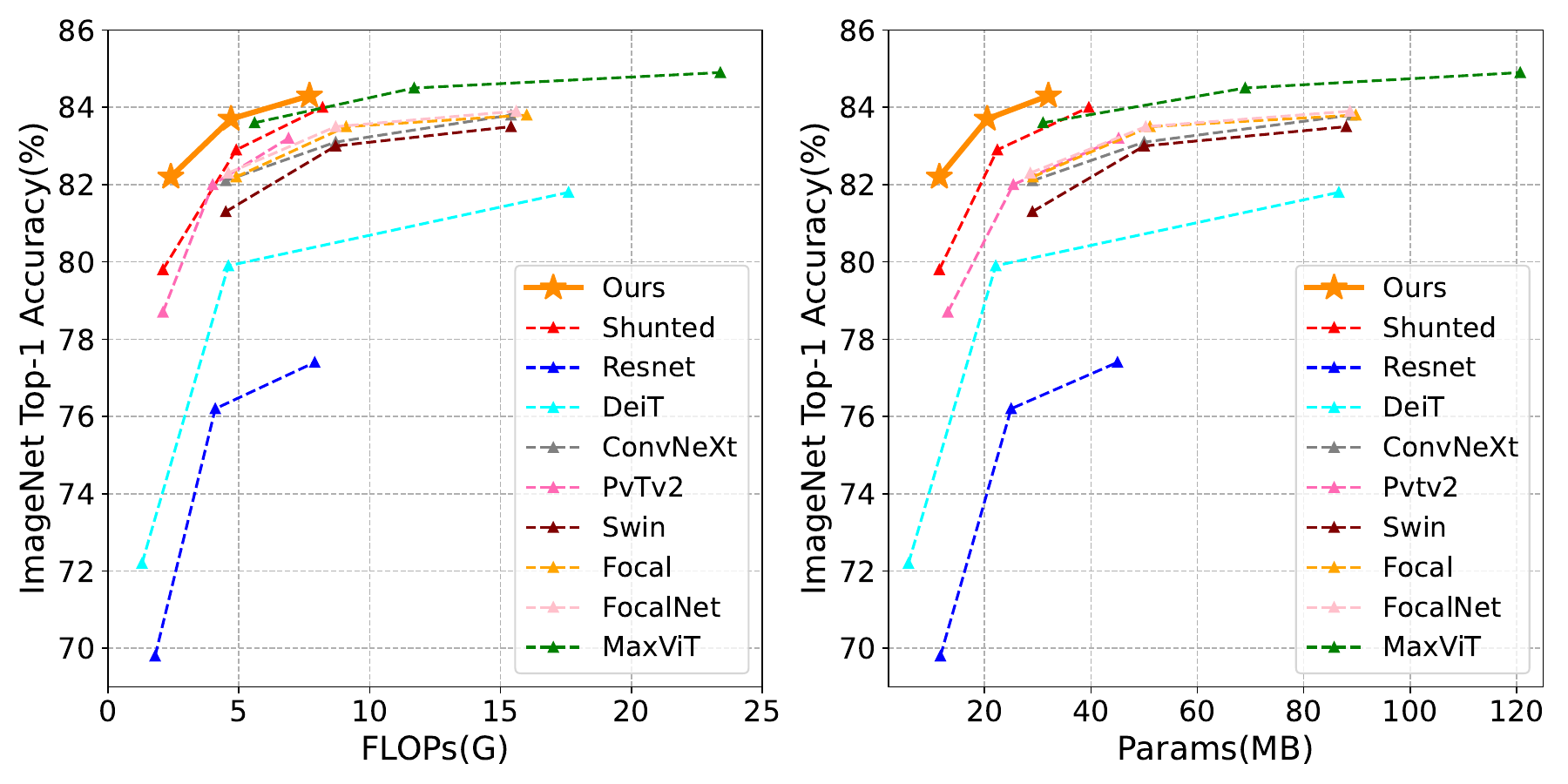}
    \caption{Top-1 accuracy on ImageNet-1K of recent SOTA models. Our proposed SMT outperforms all the baselines. 
    }
  \label{fig:sota}
\end{figure}

\section{Introduction}
\label{sec:intro}
Since the groundbreaking work on Vision Transformers (ViT)~\cite{vit}, Transformers have gained significant attention from both industry and academia, achieving remarkable success in various computer vision tasks, such as image classification~\cite{imagenet1k}, object detection~\cite{coco,pascal}, and semantic segmentation~\cite{ade20k,cityscapes}. 
Unlike convolutional networks, which only allow for interactions within a local region using a shared kernel, ViT divides the input image into a sequence of patches and updates token features via self-attention (SA), enabling global interactions. 
However, self-attention still faces challenges in downstream tasks due to the quadratic complexity in the number of visual tokens, particularly for high-resolution inputs.

To address these challenges, several efficient spatial attention techniques have been proposed. For example, Swin Transformer~\cite{swin} employs window attention to limit the number of tokens and establish cross-window connections via shifting. PVT~\cite{pvt, pvtv2} and Focal~\cite{focal} reduce the cost of self-attention by combining token merging with spatial reduction. Shunted~\cite{shunted} effectively models objects at multiple scales simultaneously while performing spatial reduction. Other techniques such as dynamic token selection~\cite{meng2022adavit,rao2021dynamicvit,yin2022vit} have also proven to be effective improvements.

 Rather than directly improving self-attention, several works~\cite{CoAtNet, efficientformer, mehta2021mobilevit, nextvit} have investigated hybrid CNN-Transformer architectures that combine efficient convolutional blocks with powerful Transformer blocks.
 We observed that most hybrid networks replace shallow Transformer blocks with convolution blocks to reduce the high computational cost of self-attention in the early stages. However, these simplistic stacking strategies hinder them from achieving a better balance between accuracy and latency. Therefore, one of the objectives of this paper is to present a new perspective on the integration of Transformer and convolution blocks.

 Based on the research conducted in~\cite{vit, uvit}, which performed a quantitative analysis of different depths of self-attention blocks and discovered that shallow blocks tend to capture short-range dependencies while deeper ones capture long-range dependencies, we propose that substituting convolution blocks for Transformer blocks in shallow networks offers a promising strategy for two primary reasons:
$(1)$ self-attention induces significant computational costs in shallow networks due to high-resolution input, and $(2)$ convolution blocks, which inherently possess a capacity for local modeling, are more proficient at capturing short-range dependencies than SA blocks in shallow networks. However, we observed that simply applying the convolution directly to the feature map does not lead to the desired performance. 
Taking inspiration from recent convolutional modulation networks~\cite{guo2022visual, conv2former, focalnet}, 
we discovered that convolutional modulation can aggregate surrounding contexts and adaptively self-modulate, giving it a stronger modeling capability than using convolution blocks alone. Therefore, we proposed a novel convolutional modulation, termed Scale-Aware Modulation (SAM), 
which incorporates two new modules: Multi-Head Mixed Convolution (MHMC) and Scale-Aware Aggregation (SAA). The MHMC module is designed to enhance the receptive field and capture multi-scale features simultaneously. The SAA module is designed to effectively aggregate features across different heads while maintaining a lightweight architecture. Despite these improvements, we find that SAM falls short of the self-attention mechanism in capturing long-range dependencies. To address this, we propose a new hybrid Modulation-Transformer architecture called the Evolutionary Hybrid Network (EHN). Specifically, we incorporate SAM blocks in the top two stages and Transformer blocks in the last two stages, while introducing a new stacking strategy in the penultimate stage. This architecture not only simulates changes in long-range dependencies from shallow to deep layers but also enables each block in each stage to better match its computational characteristics, leading to improved performance on various downstream tasks. Collectively, we refer to our proposed architecture as Scale-Aware Modulation Transformer (SMT).

As shown in Fig.~\ref{fig:sota}, our SMT significantly outperforms other SOTA vision Transformers and convolutional networks on ImageNet-1K~\cite{imagenet1k}. 
It is worth noting that our SMT achieves top-1 accuracy of 82.2\% and 84.3\% with the tiny and base model sizes, respectively. 
Moreover, our SMT consistently outperforms other SOTA models on COCO~\cite{coco} and ADE20K~\cite{ade20k} for object detection, instance segmentation, and semantic segmentation tasks.

Overall, the contributions of this paper are as follows. 
\vspace{-1mm}
\begin{itemize}
\setlength\itemsep{0em}
    \item We introduce the Scale-Aware Modulation (SAM) which incorporates a potent Multi-Head Mixed Convolution (MHMC) and an innovative, lightweight Scale-Aware Aggregation (SAA). The SAM facilitates the integration of multi-scale contexts and enables adaptive modulation of tokens to achieve more precise predictions.
    \item We propose a new evolutionary hybrid network that effectively models the transition from capturing local to global dependencies as the network increases in depth, leading to improved performance and high efficiency.
    \item We evaluated our proposed Scale-Aware Modulation Transformer (SMT) on several widely used benchmarks, including classification, object detection, and segmentation. The experimental results indicated that SMT consistently outperformed the SOTA Vision Transformers while requiring fewer parameters and incurring lower computational costs.
\end{itemize}


\begin{figure*}[t]
    \centering
        \includegraphics[width=1.0\textwidth]{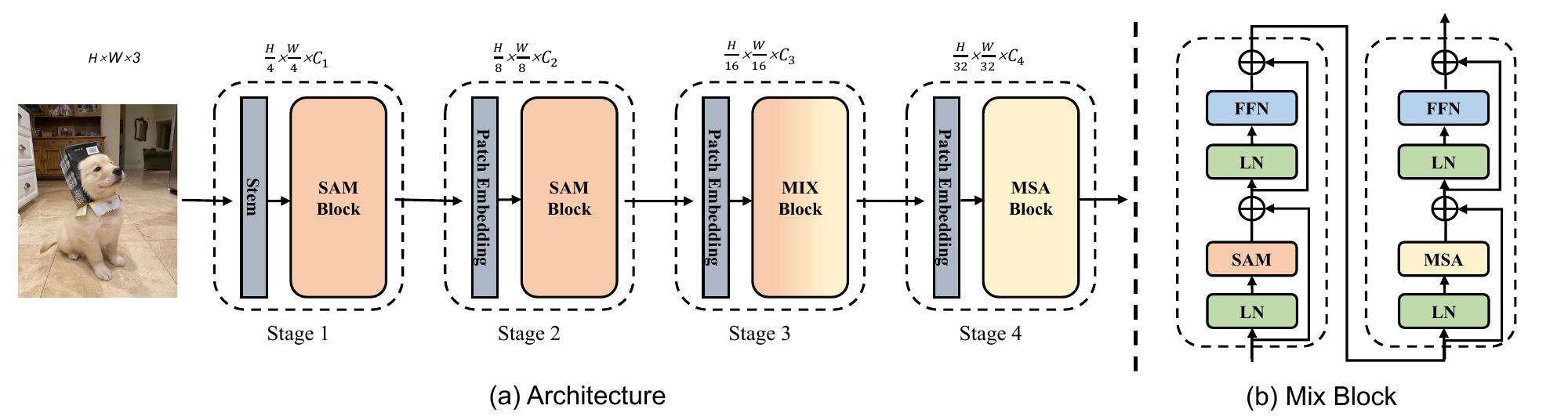}
    \caption{(a) The architecture of the Scale-Aware Modulation Transformer (SMT); (b) Mix Block: a series of SAM blocks and MSA blocks that are stacked successively (as presented in Sec.~\ref{EHS}). SAM and MSA denote the scale-aware modulation module and multi-head self-attention module, respectively.}
  \label{fig:overview}\vspace{-2mm}
\end{figure*}

\section{Related Work}

\subsection{Vision Transformers}
The Transformer~\cite{vaswani2017attention} was initially developed for natural language processing tasks and has since been adapted for computer vision tasks through the introduction of the Vision Transformer (ViT)~\cite{vit}. Further improvements to ViT have been achieved through knowledge distillation or more intricate data augmentation, as demonstrated by DeiT~\cite{deit}. However, Transformers do not consider the quadratic complexity of high-resolution images or the 2D structure of images, which are challenges in vision tasks. To address these issues and improve the performance of vision Transformers, various methods have been proposed, including multi-scale architectures~\cite{crossvit, swin, pvt, xu2021co}, lightweight convolution layers~\cite{cmt, localvit, cvt}, and local self-attention mechanisms~\cite{swin, twins, focal, zhang2021multi}. 

\subsection{Convolutional Neural Networks}
Convolutional neural networks (CNNs) have been the main force behind the revival of deep neural networks in computer vision. 
Since the introduction of AlexNet~\cite{krizhevsky2017imagenet}, VGGNet~\cite{simonyan2014very}, and ResNet~\cite{resnet}, CNNs have rapidly become the standard framework for computer vision tasks. The design principles of CNNs have been advanced by subsequent models such as Inception~\cite{inceptionv4, szegedy2016rethinking}, ResNeXt~\cite{xie2017aggregated}, Res2Net~\cite{pami21Res2net} and MixNet~\cite{tan2019mixconv}, which promote the use of building blocks with multiple parallel convolutional paths. Other works such as MobileNet~\cite{mobilenet} and ShuffleNet~\cite{shufflenet} have focused on the efficiency of CNNs. To further improve the performance of CNNs, attention-based models such as SE-Net~\cite{senet}, Non-local Networks~\cite{wang2018non}, and CBAM~\cite{cbam} have been proposed to enhance the modeling of channel or spatial attention. EfficientNets~\cite{efficientnet, efficientnetv2} and MobileNetV3~\cite{mobilenetv3} have employed neural architecture search (NAS)~\cite{nas} to develop efficient network architectures. ConvNeXt~\cite{convnext} adopts the hierarchical design of Vision Transformers to enhance CNN performance while retaining the simplicity and effectiveness of CNNs. 
Recently, several studies~\cite{guo2022visual, conv2former, focalnet} have utilized convolutional modulation as a replacement for self-attention, resulting in improved performance.
Specifically, FocalNet~\cite{focalnet} utilizes a stack of depth-wise convolutional layers to encode features across short to long ranges and then injects the modulator into the tokens using an element-wise affine transformation. Conv2Former~\cite{conv2former} achieves good recognition performance using a simple $11\times11$ depth-wise convolution. In contrast, our scale-aware modulation also employs depth-wise convolution as a basic operation but introduces multi-head mixed convolution and scale-aware aggregation. 

\subsection{Hybrid CNN-Transformer Networks}
A popular topic in visual recognition is the development of hybrid CNN-Transformer architectures. Recently, several studies~\cite{cmt, BoTNet, cvt, FAN} have demonstrated the effectiveness of combining Transformers and convolutions to leverage the strengths of both architectures. CvT~\cite{cvt} first introduced depth-wise and point-wise convolutions before self-attention. CMT~\cite{cmt} proposed a hybrid network that utilizes Transformers to capture long-range dependencies and CNNs to model local features. 
MobileViT~\cite{mehta2021mobilevit}, EdgeNeXt~\cite{edgenext}, MobileFormer~\cite{mobileformer}, and EfficientFormer~\cite{efficientformer} reintroduced convolutions to Transformers for efficient network design and demonstrated exceptional performance in image classification and downstream applications. However, the current hybrid networks lack the ability to model range dependency transitions, making it challenging to improve their performance. In this paper, we propose an evolutionary hybrid network that addresses this limitation and showcases its importance.

\section{Method}
\subsection{Overall Architecture}
The overall architecture of our proposed Scale-Aware Modulation Transformer (SMT) is illustrated in Fig.~\ref{fig:overview}. The network comprises four stages, each with downsampling rates of $\{4,8,16,32\}$.
Instead of constructing an attention-free network, we first adopt our proposed Scale-Aware Modulation (SAM) in the top two stages, followed by a penultimate stage where we sequentially stack one SAM block and one Multi-Head Self-Attention (MSA) block to model the transition from capturing local to global dependencies. For the last stage, we solely use MSA blocks to capture long-range dependencies effectively. For the Feed-Forward Network (FFN) in each block, we adopt the detail-specific feedforward layers as used in Shunted~\cite{shunted}. 

\subsection{Scale-Aware Modulation}
\begin{figure}[t]
    \centering
        \includegraphics[width=0.49\textwidth]{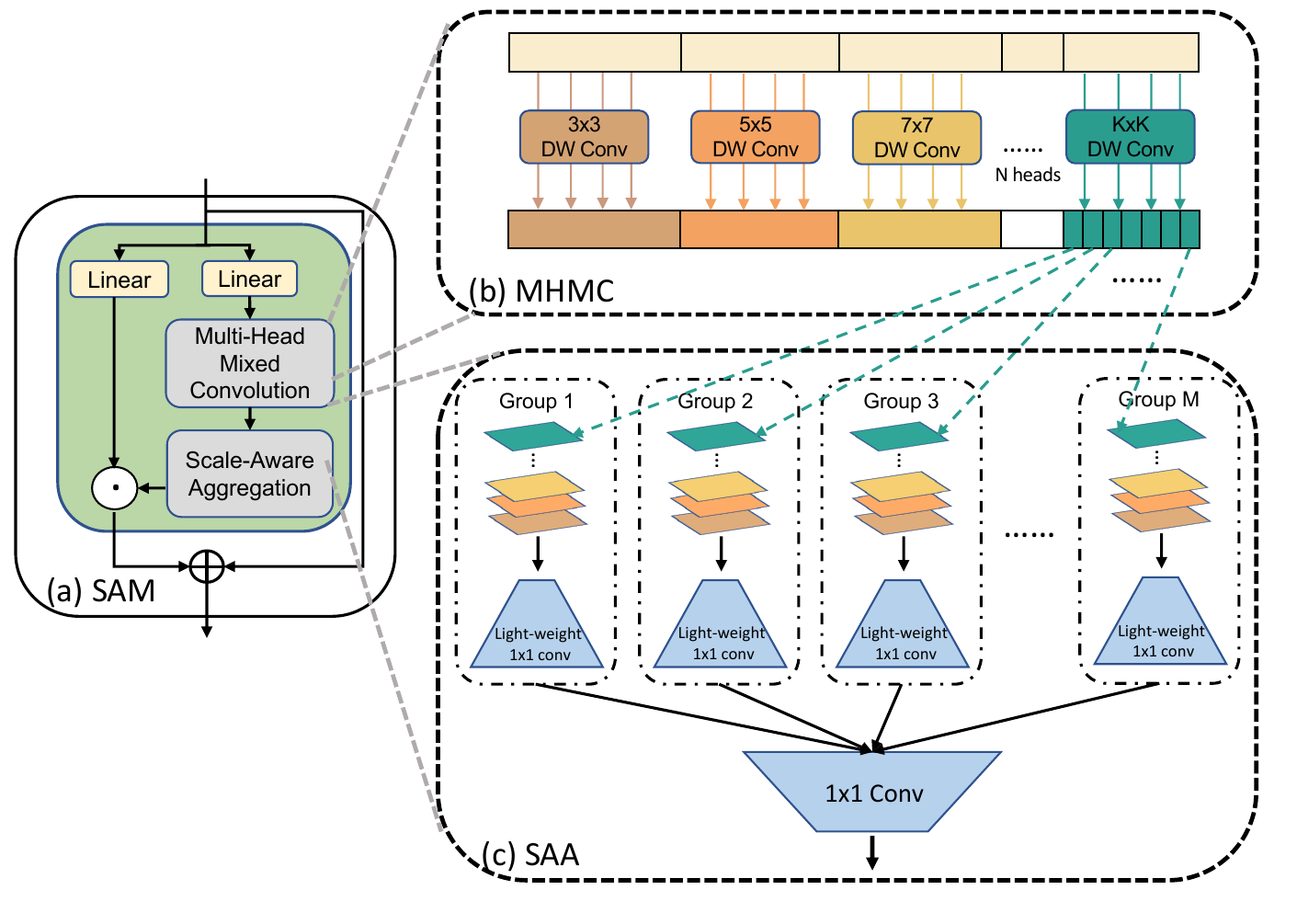}
    \caption{(a) The schematic illustration of the proposed scale-aware modulation (SAM). (b) and (c) are the module descriptions of multi-head mixed convolution (MHMC) and scale-aware aggregation (SAA), respectively.
}
  \label{fig:sam}\vspace{-2mm}
\end{figure}

\paragraph{Multi-Head Mixed Convolution} 

We propose the Multi-Head Mixed Convolution (MHMC), which introduces multiple convolutions with different kernel sizes, enabling it to capture various spatial features across multiple scales. Furthermore, MHMC can expand the receptive field using a large convolutional kernel, enhancing its ability to model long-range dependencies. As depicted in Fig.~\ref{fig:sam}(b), MHMC partitions input channels into N heads and applies distinct depth-wise separable convolutions to each head, which reduces the parameter size and computational cost. To simplify our design process, we initialize the kernel size with 3$\times$3 and gradually increase it by 2 per head. This approach enables us to regulate the range of receptive fields and multi-granularity information by merely adjusting the number of heads.
Our proposed MHMC can be formulated as follows:
\begin{equation}
\begin{aligned}
MHMC(X)&= Concat(DW_{k_1\times k_1}(x_1), \dots, DW_{k_n \times k_n}(x_n)) \\
\end{aligned}
\end{equation}
where $x = [x_1, x_2, ... , x_n]$ means to split up the input feature $x$ into multiple heads in the channel dimension and $k_i \in \{3, 5, \dots, K\}$ denotes the kernel size increases monotonically by 2 per head.

\begin{figure}[h]
    \centering
        \includegraphics[width=0.48\textwidth]{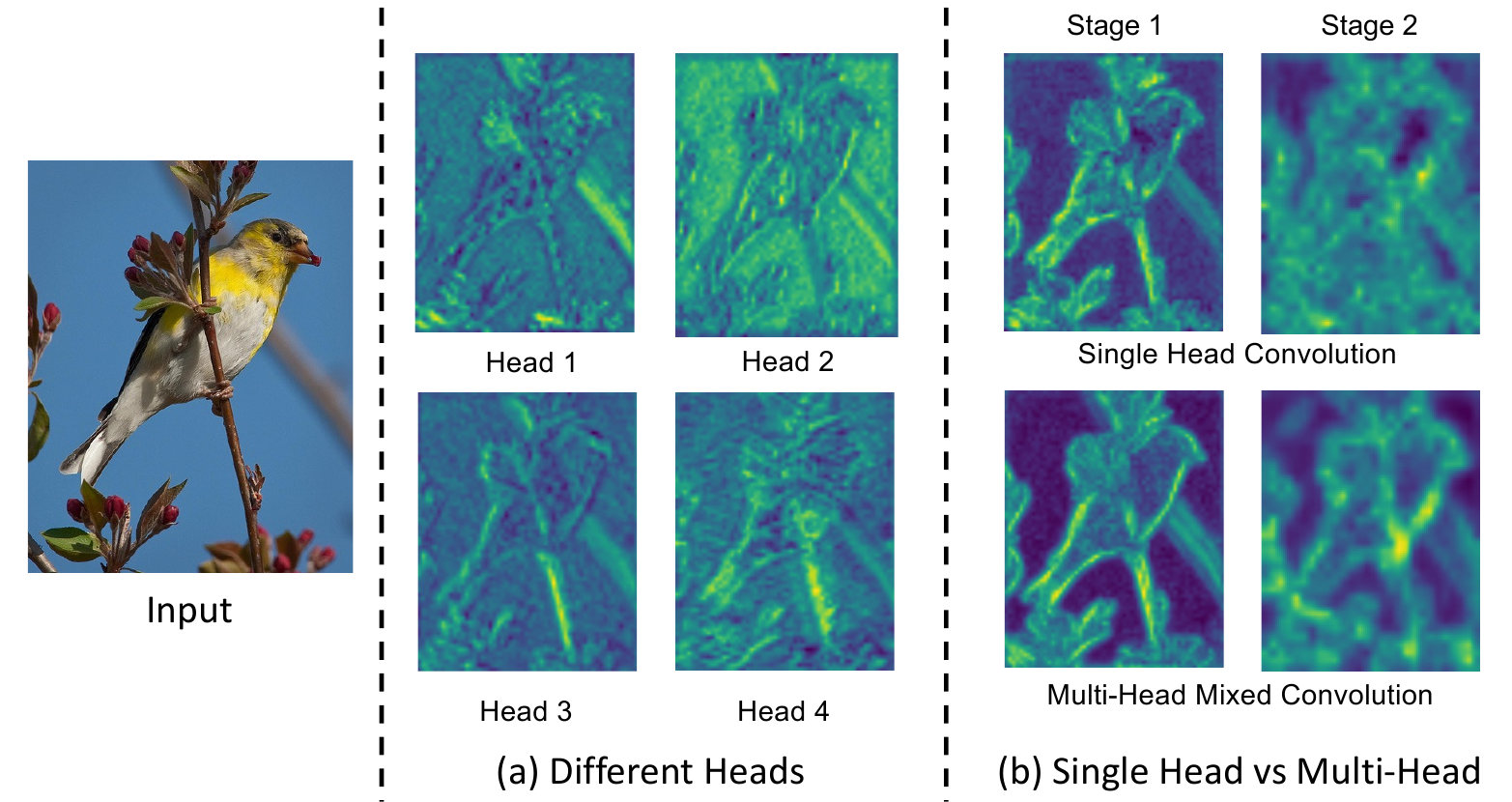}
    \caption{(a) Visualization of the output values of different heads in the MHMC in the first stage. (b) Visualization of the modulation values (corresponding to the left side of $\odot$ in Eq.~\ref{formula:modulation}) under single-head and multi-head mixed convolution in the last layer during the top two stages. All maps are upsampled for display.}
  \label{fig:vis_heads}
\end{figure}

As shown in Fig.~\ref{fig:vis_heads}(a), each distinct convolution feature map learns to focus on different granularity features in an adaptive manner, as expected. Notably, when we compare the single-head and multi-head by visualizing modulation maps in Fig.~\ref{fig:vis_heads}(b), we find that the visualization under multi-head depicts the foreground and target objects accurately in stage 1, while filtering out background information effectively. Moreover, it can still present the overall shape of the target object as the network becomes deeper, while the information related to the details is lost under the single-head convolution. This indicates that MHMC has the ability to capture local details better than a single head at the shallow stage, while maintaining detailed and semantic information about the target object as the network becomes deeper.

\paragraph{Scale-Aware Aggregation} 

To enhance information interaction across multiple heads in MHMC, we introduce a new lightweight aggregation module, termed Scale-Aware Aggregation (SAA), as shown in Fig.~\ref{fig:sam}(c). The SAA involves an operation that shuffles and groups the features of different granularities produced by the MHMC. Specifically, we select one channel from each head to construct a group, and then we utilize the inverse bottleneck structure to perform an up-down feature fusion operation within each group, thereby enhancing the diversity of multi-scale features. However, a well-designed grouping strategy enables us to introduce only a small amount of computation while achieving desirable aggregation results. Notably, let the input $X \in \mathbb{R}^{H\times W\times C}$, $ Groups = \frac{C}{Heads} $, which means the number of groups is inversely proportional to the number of heads. Subsequently, we perform cross-group information aggregation for all features using point-wise convolution to achieve cross-fertilization of global information.
The process of SAA can be formulated as follows:
\begin{equation}
\begin{aligned}
M &= W_{inter}([G_1,G_2, \dots, G_M]), \\ 
G_i &= W_{intra}([H^i_1,H^i_2, \dots, H^i_N]), \\
H^i_j &= DWConv_{k_j\times k_j}(x^i_j) \in \mathbb{R}^{H\times W\times 1}. \\
\end{aligned}
\end{equation}
where $W_{inter}$ and $W_{intra}$ are weight matrices of point-wise convolution. $j \in \{1, 2, \dots, N\}$ and $i \in \{1, 2, \dots, M\}$, where $N$ and $M = \frac{C}{N}$ denote the number of heads and groups, respectively. Here, $H_j \in \mathbb{R}^{H\times W\times M}$ represents the $j$-th head with depth-wise convolution, and $H^i_j$ represents the $i$-th channel in the $j$-th head.

\begin{figure}[h]
    \centering
        \includegraphics[width=0.48\textwidth]{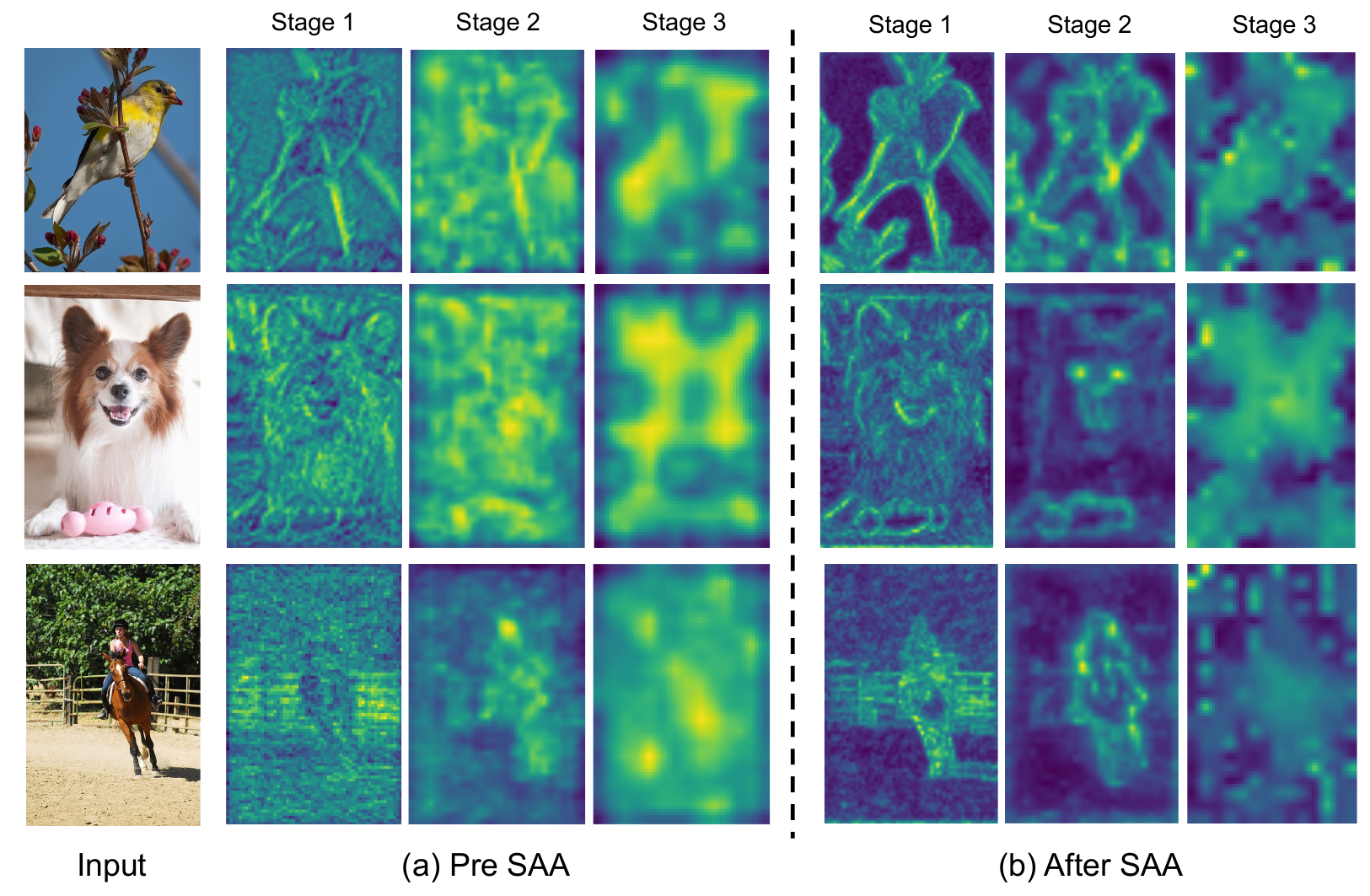}
    \caption{(a) Visualization of the modulation values before SAA. (b) Visualization of the modulation values after SAA.}
  \label{fig:vis_aggregation}\vspace{-2mm}
\end{figure}

Fig.~\ref{fig:vis_aggregation} shows that our SAA module explicitly strengthens the semantically relevant low-frequency signals and precisely focuses on the most important parts of the target object. For instance, in stage 2, the eyes, head and body are clearly highlighted as essential features of the target object, resulting in significant improvements in classification performance. Compared to the convolution maps before aggregation, our SAA module demonstrates a better ability to capture and represent essential features for visual recognition tasks. (More visualizations can be found in Appendix~\ref{Appendix:E}).

\paragraph{Scale-Aware Modulation} 
As illustrated in Fig.~\ref{fig:sam}(a), after capturing multi-scale spatial features using MHMC and aggregating them with SAA, we obtain an output feature map, which we refer to as the modulator M. We then adopt this modulator to modulate the value V using the scalar product. For the input features $X \in \mathbb{R}^{H\times W\times C}$, we compute the output Z as follows:

\begin{equation}
\label{formula:modulation}
\begin{aligned}
    Z &= M \odot V, \\
    V &= W_v X,  \\
    M &= SAA(MHMC(W_s X)).
\end{aligned}
\end{equation}
where $\odot$ is the element-wise multiplication, $W_v$ and $W_s$ are weight martices of linear layers. 
Since the modulator is calculated via Eq.~\ref{formula:modulation}, it changes dynamically with different inputs, thereby achieving adaptively self-modulation. Moreover, unlike self-attention, which computes an $N \times N$ attention map, the modulator retains the channel dimension. This feature allows for spatial- and channel-specific modulation of the value after element-wise multiplication, while also being memory-efficient, particularly when processing high-resolution images.

\subsection{Scale-Aware Modulation Transformer}
\begin{figure}[t]
    \centering
        \includegraphics[width=0.44\textwidth]{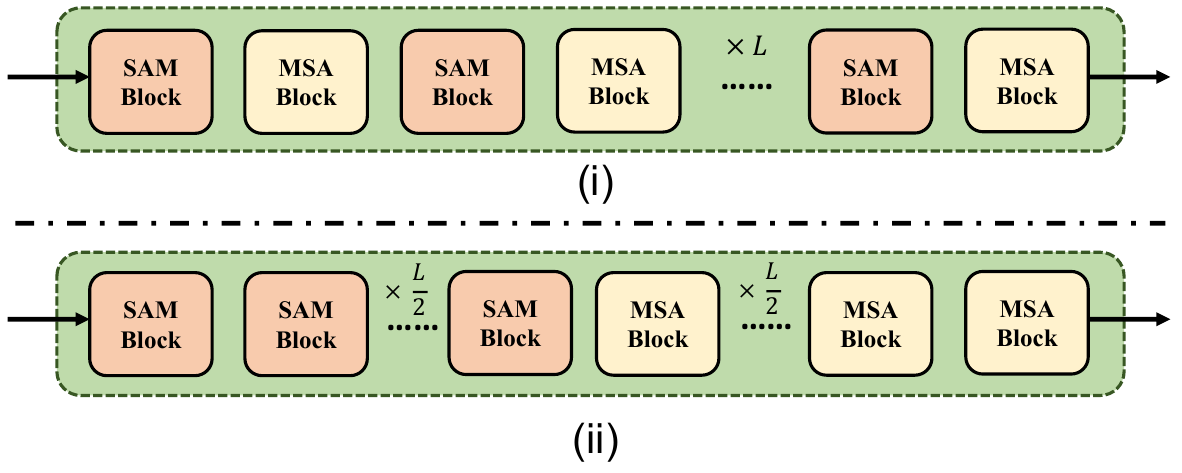}
    \caption{Two proposed hybrid stacking strategies.}
  \label{fig:hybrid}\vspace{-2mm}
\end{figure}

\paragraph{Evolutionary Hybrid Network}
\label{EHS}
In this section, we propose to reallocate the appropriate computational modules according to the variation pattern in the network's capture range dependencies to achieve better computational performance.
We propose using MSA blocks only from the penultimate stage to reduce the computational burden.
Furthermore, to effectively simulate the transition pattern, we put forth two hybrid stacking strategies for the penultimate stage: $(i)$  sequentially stacking one SAM block and one MSA block, which can be formulated as $(SAM \times 1 + MSA \times 1) \times \frac{N}{2}$, depicted in Fig.~\ref{fig:hybrid}(i); $(ii)$ using SAM blocks for the first half of the stage and MSA blocks for the second half, which can be formulated as $(SAM \times \frac{N}{2} + MSA \times \frac{N}{2})$, depicted in Fig.~\ref{fig:hybrid}(ii).

\begin{figure}[h]
    \centering
        \includegraphics[width=0.35\textwidth]{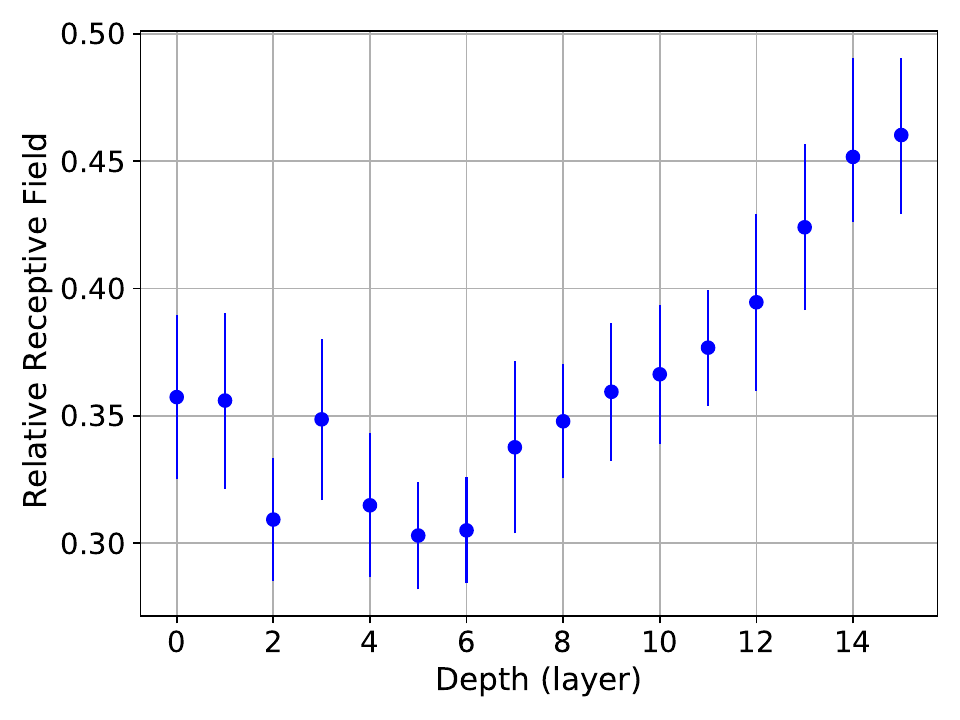}
    \caption{The receptive field of SMT-B's relative attention across depth, with error bars representing standard deviations across various attention heads.}
  \label{fig:attn_distance}\vspace{-2mm}
\end{figure}

To assess the efficacy of these hybrid stacking strategies, we evaluated their top-1 accuracy on the ImageNet-1K, as shown in Table~\ref{tab:ehs}. Moreover, as depicted in Fig.~\ref{fig:attn_distance}, we calculate the relative receptive field of the MSA blocks in the penultimate stage, followed by the approach presented in~\cite{uvit}. It is noteworthy that there is a slight downward trend in the onset of the relative receptive field in the early layers. This decline can be attributed to the impact of the SAM on the early MSA blocks, which emphasize neighboring tokens. We refer to this phenomenon as the adaptation period. As the network becomes deeper, we can see a smooth and steady upward trend in the receptive field, indicating that our proposed evolutionary hybrid network effectively simulates the transition from local to global dependency capture.

\section{Experiments}

\begin{table}[t]
\setlength\tabcolsep{2.5pt}
\centering
\small
\begin{tabular}{c|ccc|c}
\Xhline{1.0pt}
\multicolumn{5}{c}{\textbf{(a) Tiny Models}} \\
method &  \begin{tabular}[c]{@{}c@{}}image \\ size\end{tabular} & \#param. & FLOPs & \begin{tabular}[c]{@{}c@{}}ImageNet \\ top-1 acc.\end{tabular} \\
\hline
RegNetY-1.6G~\cite{regnet} & 224$^2$ & 11.2M & 1.6G & 78.0 \\
EffNet-B3~\cite{efficientnet} & 300$^2$ & 12M & 1.8G & 81.6 \\
PVTv2-b1~\cite{pvtv2} & 224$^2$ & 13.1M & 2.1G & 78.7 \\ 
EfficientFormer-L1~\cite{efficientformer} & 224$^2$ & 12.3M & 1.3G & 79.2 \\ 
Shunted-T~\cite{shunted} & 224$^2$ & 11.5M & 2.1G & 79.8 \\ 
Conv2Former-N~\cite{conv2former} & 224$^2$ & 15M & 2.2G & 81.5 \\ 
\hline
\textbf{SMT-T(Ours)} & 224$^2$ & 11.5M & 2.4G & \textbf{82.2} \\

\Xhline{1.0pt}
\multicolumn{5}{c}{\textbf{(b) Small Models}}  \\  
method & \begin{tabular}[c]{@{}c@{}}image \\ size\end{tabular} & \#param. & FLOPs & \begin{tabular}[c]{@{}c@{}}ImageNet \\ top-1 acc.\end{tabular} \\
\hline
RegNetY-4G~\cite{regnet} & 224$^2$ & 21M & 4.0G & 80.0 \\
EffNet-B4~\cite{efficientnet} & 380$^2$ & 19M & 4.2G & 82.9 \\ 
DeiT-S~\cite{deit} & 224$^2$ & 22M & 4.6G & 79.8 \\
Swin-T~\cite{swin} & 224$^2$ & 29M & 4.5G &  81.3 \\
ConvNeXt-T ~\cite{convnext} & 224$^2$ & 29M & 4.5G & 82.1 \\
PVTv2-b2 ~\cite{pvtv2} & 224$^2$ & 25.0M & 4.0G & 82.0 \\
Focal-T ~\cite{focal} & 224$^2$ & 29.1M & 4.9G &  82.2 \\
Shunted-S~\cite{shunted} & 224$^2$ & 22.4M & 4.9G &  82.9 \\
CMT-S~\cite{cmt} & 224$^2$ & 25.1M & 4.0G & 83.5 \\ 
FocalNet-T~\cite{focalnet} & 224$^2$ & 28.6M & 4.5G &  82.3 \\
Conv2Former-T~\cite{conv2former} & 224$^2$ & 27M & 4.4G &  83.2 \\ 
HorNet-T~\cite{hornet} & 224$^2$ & 23M & 4.0G & 83.0 \\
InternImage-T~\cite{internimage} & 224$^2$ & 30M & 5.0G &  83.5 \\ 
MaxViT-T~\cite{tu2022maxvit} & 224$^2$ & 31M & 5.6G &83.6 \\
\hline
\textbf{SMT-S(Ours)} & 224$^2$ & 20.5M & 4.7G & \textbf{83.7} \\

\Xhline{1.0pt}
\multicolumn{5}{c}{\textbf{(c) Base Models}}  \\  
method & \begin{tabular}[c]{@{}c@{}}image \\ size\end{tabular} & \#param. & FLOPs & \begin{tabular}[c]{@{}c@{}}ImageNet \\ top-1 acc.\end{tabular} \\
\hline
RegNetY-8G~\cite{regnet} & 224$^2$ & 39M & 8.0G & 81.7 \\
EffNet-B5 ~\cite{efficientnet} & 456$^2$ & 30M & 9.9G &  83.6 \\
Swin-S~\cite{swin} & 224$^2$ & 49.6M & 8.7G & 83.0 \\
CoAtNet-1~\cite{CoAtNet} & 224$^2$ & 42M & 8.0G & 83.3 \\
PVTv2-b4~\cite{pvtv2} & 224$^2$ & 63M & 10.0G & 83.6 \\
SwinV2-S/8~\cite{swinv2} & 256$^2$ & 50M & 12.0G & 83.7 \\
PoolFormer-m36~\cite{poolformer} & 224$^2$ & 56.2M & 8.8G & 82.1 \\

Shunted-B~\cite{shunted} & 224$^2$ & 39.6M & 8.1G & 84.0 \\
InternImage-S~\cite{internimage} & 224$^2$ & 50.0M & 8.0G &  84.2 \\ 
Conv2Former-S~\cite{conv2former} & 224$^2$ & 50.0M & 8.7G &  84.1 \\ 
Swin-B~\cite{swin} & 224$^2$ & 87.8M & 15.4G &  83.4 \\
ConvNeXt-B ~\cite{convnext} & 224$^2$ & 89M & 15.4G & 83.8 \\
Focal-B~\cite{focal} & 224$^2$ & 89.8M & 16.4G & 83.8 \\
FocalNet-B~\cite{focalnet} & 224$^2$ & 88.7M & 15.4G & 83.9 \\
HorNet-B~\cite{hornet} & 224$^2$ & 87M & 15.6G & 84.2 \\
\hline
\textbf{SMT-B(Ours)} & 224$^2$ & 32.0M & 7.7G & \textbf{84.3} \\

\Xhline{1.0pt}
\end{tabular}
\normalsize
\caption{Comparison of different backbones on ImageNet-1K classification. 
}
\label{exp:imagenet}\vspace{-2mm}
\end{table}

\begin{table}[h]
\setlength\tabcolsep{2.0pt}
\centering
\small
\begin{tabular}{c|ccc|c}
\Xhline{1.0pt}
\multicolumn{5}{c}{\textbf{ImageNet-22K pre-trained models}} \\
method &  \begin{tabular}[c]{@{}c@{}}image \\ size\end{tabular} & \#param. & FLOPs & \begin{tabular}[c]{@{}c@{}}ImageNet \\ top-1 acc.\end{tabular} \\
\hline
ViT-B/16~\cite{vit} & 384$^2$ & 86.0M & 55.4G & 84.0 \\
ViT-L/16~\cite{vit} & 384$^2$ & 307.0M & 190.7G & 85.2 \\
\hline
Swin-Large~\cite{swin} & 224$^2$/224$^2$ & 196.5M & 34.5G & 86.3 \\ 
Swin-Large~\cite{swin} & 384$^2$/384$^2$ & 196.5M & 104.0G & 87.3 \\ 
\hline
FocalNet-Large~\cite{focalnet} & 224$^2$/224$^2$ & 197.1M & 34.2G & 86.5 \\ 
FocalNet-Large~\cite{focalnet} & 224$^2$/384$^2$ & 197.1M & 100.6G & 87.3 \\ 
\hline
InternImage-L~\cite{internimage} & 224$^2$/384$^2$ & 223M & 108G & 87.7 \\ 
InternImage-XL~\cite{internimage} & 224$^2$/384$^2$ & 335M & 163G & 88.0 \\ 
\hline
\textbf{SMT-L(Ours)} & 224$^2$/224$^2$ & 80.5M & 17.7G & \textbf{87.1} \\
\textbf{SMT-L(Ours)} & 224$^2$/384$^2$ & 80.5M & 54.6G & \textbf{88.1} \\

\Xhline{1.0pt}
\end{tabular}
\normalsize
\caption{ImageNet-1K finetuning results with models pretrained on ImageNet-22K. Numbers before and
after “/” are resolutions used for pretraining and finetuning, respectively}
\label{imagenet22k}\vspace{-4mm}
\end{table}

To ensure a fair comparison under similar parameters and computation costs, we construct a range of SMT variants.
We validate our SMTs on ImageNet-1K~\cite{imagenet1k} image classification, MS COCO~\cite{coco} object detection, and ADE20K~\cite{ade20k} semantic segmentation. Besides, extensive ablation studies provide a close look at different components of the SMT. (The detailed model settings are presented in Appendix~\ref{Appendix:A})

\begin{table*}[t]
\setlength\tabcolsep{2.pt}
\centering
\small
{
\begin{tabular}{l|c|c|cccccc|cccccc}
\toprule
\multirow{2}{*}{Backbone} & Params & FLOPs&  \multicolumn{6}{c|}{Mask R-CNN 1$\times$ schedule} & \multicolumn{6}{c}{Mask R-CNN 3$\times$ schedule + MS} \\

                          &   (M)   &   (G)    &   $AP^b$    &  $AP^b_{50}$     &  $AP^b_{75}$     &   $AP^m$    &   $AP^m_{50}$    &  $AP^m_{75}$     &    $AP^b$    &  $AP^b_{50}$     &  $AP^b_{75}$     &   $AP^m$    &   $AP^m_{50}$    &  $AP^m_{75}$     \\
\midrule
ResNet50~\cite{resnet} &   44.2  & 260   &   38.0& 58.6 &41.4 &34.4& 55.1& 36.7 &41.0& 61.7& 44.9& 37.1& 58.4& 40.1 \\
Twins-SVT-S~\cite{twins} & 44.0 & 228    &  43.4 &66.0& 47.3 &40.3 &63.2& 43.4 &46.8& 69.2 &51.2 &42.6& 66.3 &45.8\\
Swin-T~\cite{swin} & 47.8 & 264& 42.2 &64.6& 46.2 &39.1 &61.6& 42.0 &46.0& 68.2 &50.2 &41.6& 65.1 &44.8\\
PVTv2-B2~\cite{pvt}    &    45.0   & -   & 45.3& 67.1& 49.6& 41.2& 64.2& 44.4& -&-&-&-&-&-\\
Focal-T~\cite{focal} & 48.8 & 291 & 44.8 &67.7& 49.2& 41.0& 64.7& 44.2 & 47.2 &69.4 &51.9& 42.7& 66.5& 45.9 \\
CMT-S~\cite{cmt} & 44.5 & 249 & 44.6 & 66.8 & 48.9 & 40.7 & 63.9 & 43.4 & -&-&-&-&-&- \\
FocalNet-T~\cite{focalnet} & 48.9 & 268   & 46.1& 68.2& 50.6& 41.5 &65.1 &44.5& 48.0& 69.7 & 53.0 & 42.9 & 66.5 & 46.1 \\
\rowcolor{mygray}
\textbf{SMT-S} & \textbf{40.0}& 265 & \textbf{47.8}& \textbf{69.5}& \textbf{52.1} &\textbf{43.0}& \textbf{66.6} &\textbf{46.1} &\textbf{49.0} &\textbf{70.1} &\textbf{53.4} &\textbf{43.4} &\textbf{67.3} &\textbf{46.7}\\
\midrule
ResNet101~\cite{resnet}  &   63.2  &   336    &  40.4& 61.1 &44.2 &36.4& 57.7& 38.8 &42.8& 63.2& 47.1& 38.5& 60.1& 41.3 \\
Swin-S~\cite{swin} & 69.1 &  354 & 44.8 &66.6& 48.9 &40.9 &63.4& 44.2 &48.5& 70.2 &53.5 &43.3& 67.3 &46.6\\
Swin-B~\cite{swin} & 107.1 &  497 & 46.9 & 69.2 & 51.6 & 42.3 & 66.0 & 45.5 &48.5& 69.8 &53.2 &43.4& 66.8 &46.9\\
Twins-SVT-B~\cite{twins} & 76.3 &  340  & 45.2 &67.6& 49.3 &41.5 &64.5& 44.8 &48.0& 69.5 &52.7 &43.0&66.8 &46.6\\
PVTv2-B4~\cite{pvt}    &    82.2   &   -  & 47.5& 68.7& 52.0& 42.7& 66.1& 46.1& -&-&-&-&-&- \\
Focal-S~\cite{focal} & 71.2 &  401  & 47.4 &69.8 & 51.9& 42.8& 66.6& 46.1 & 48.8 &70.5 &53.6& 43.8& 67.7& 47.2 \\
FocalNet-S~\cite{focalnet} & 72.3 & 365   & 48.3& \textbf{70.5}& 53.1& 43.1 &67.4 &46.2& 49.3& 70.7 & 54.2 & 43.8 & 67.9 & 47.4 \\

\rowcolor{mygray}
\textbf{SMT-B} & \textbf{51.7} & \textbf{328} & \textbf{49.0} &70.2 &\textbf{53.7} &\textbf{44.0} & \textbf{67.6}& \textbf{47.4} & \textbf{49.8} &\textbf{71.0} &\textbf{54.4} & \textbf{44.0}& \textbf{68.0}& \textbf{47.3} \\
\bottomrule
\end{tabular}}
\caption{Object detection and instance segmentation with Mask R-CNN on COCO. Only the 3$\times$ schedule has the multi-scale training. All backbones are pre-trained on ImageNet-1K. 
}
\label{tab:maskrcnn}
\end{table*}

\begin{table}[h]
\begin{center}
\resizebox{\linewidth}{!}{
\setlength{\tabcolsep}{1.4mm}{
\addtolength{\tabcolsep}{-2.5pt}
\begin{tabular}{c|c|c|c|c|c|c|c|c|c}
\Xhline{3\arrayrulewidth}
Method & Backbones & \#Params & FLOPs & $AP^b$ & $AP^b_{50}$ & $AP^b_{75}$ & $AP^m$ & $AP^m_{50}$ & $AP^m_{75}$ \\
\hline
\multirow{5}{*}{Cascade~\cite{cai2018cascade}} 
 & ResNet50~\cite{resnet} & 82.0M & 739G & 46.3 & 64.3 & 50.5 & 40.1 & 61.7 & 43.4\\
 & Swin-T~\cite{swin} & 86.0M & 745G & 50.5 & 69.3 & 54.9 & 43.7 & 66.6 & 47.1\\
 & ConvNeXt~\cite{convnext} & - & 741G & 50.4 & 69.1 & 54.8 & 43.7 & 66.5 & 47.3 \\
 & Shuffle-T~\cite{shuffleT} & 86.0M & 746G & 50.8 & 69.6 & 55.1 & 44.1 & 66.9 & 48.0\\
 & FocalNet-T~\cite{focalnet}  & 87.1M & 751G  & 51.5 & 70.3 & 56.0 & - & - & - \\
\rowcolor{mygray}
 \cellcolor{white} & \textbf{SMT-S} & \textbf{77.9M} & 744G  & \textbf{51.9} & \textbf{70.5} & \textbf{56.3} & \textbf{44.7} & \textbf{67.8} & \textbf{48.6} \\
\hline
\hline
Method & Backbones & \#Params & FLOPs & $AP^b$ & $AP^b_{50}$ & $AP^b_{75}$ & $AP_S$ & $AP_M$ & $AP_L$ \\
\hline
\multirow{5}{*}{RetinaNet~\cite{lin2017focal}} 
& ResNet50~\cite{resnet} & 37.7M & 240G & 39.0 & 58.4 & 41.8 & 22.4 & 42.8 & 51.6\\
& Swin-T~\cite{swin} & 38.5M & 245G & 45.0 & 65.9 & 48.4 & 29.7 & 48.9 & 58.1\\
& Focal-T~\cite{focal}  & 39.4M & 265G & 45.5 & 66.3 & 48.8 & 31.2 & 49.2 & 58.7\\
& Shunted-S~\cite{shunted}  & 32.1M & - & 46.4 & 66.7 & 50.4 & 31.0 & 51.0 & 60.8\\
\rowcolor{mygray}
 \cellcolor{white} & \textbf{SMT-S} & \textbf{30.1M} & 247G  & \textbf{47.3} & \textbf{67.8} & \textbf{50.5} & \textbf{32.5} & \textbf{51.1} & \textbf{62.3} \\
\Xhline{3\arrayrulewidth}
\end{tabular}
}}
\end{center}
\caption{COCO detection and segmentation with the \textbf{Cascade Mask R-CNN} and \textbf{RetinaNet}. The performances are reported on the COCO {\em val} dataset under the $3\times$ schedule.}
\label{tab:cascade}\vspace{-2mm}
\end{table}

\begin{table}[h]
\setlength{\tabcolsep}{1.0pt}
\centering
\footnotesize
  \begin{tabular}{cl|ccclll}
    \toprule
    Method & Backbone & \#Param. & FLOPs & $AP^b$ & $AP^b_{50}$ & $AP^b_{75}$ \\
    \midrule	 
    \multirow{5}{*}{Sparse R-CNN~\cite{sun2021sparse}} 
         & R-50~\cite{resnet} & 106.1M & 166G & 44.5 & 63.4 & 48.2 \\
         & Swin-T~\cite{swin} & 109.7M & 172G & 47.9 & 67.3 & 52.3 \\          
         & Focal-T~\cite{focal} & 110.8M & 196G & {49.0} & {69.1} & {53.2} \\
         & FocalNet-T~\cite{focalnet}  & 111.2M & 178G & 49.9 & 69.6 & 54.4   \\
         \rowcolor{mygray}
          \cellcolor{white} & SMT-S & \textbf{102.0M} & 171G & \textbf{50.2} & \textbf{69.8} & \textbf{54.7}   \\
    \midrule
     \multirow{5}{*}{ATSS~\cite{zhang2020bridging}} 
         & R-50~\cite{resnet} & 32.1M & 205G & 43.5 & 61.9 & 47.0 \\
         & Swin-T~\cite{swin} & 35.7M & 212G & 47.2 & 66.5 & 51.3 \\         
         & Focal-T~\cite{focal} & 36.8M & 239G & 49.5 & 68.8 & 53.9\\
         & FocalNet-T~\cite{focalnet} & 37.2M & 220G & 49.6 & 68.7 & 54.5 \\   
         \rowcolor{mygray}
          \cellcolor{white} & SMT-S & \textbf{28.0M} & 214G & \textbf{49.9} & \textbf{68.9} & \textbf{54.7}   \\
    \midrule
     \multirow{4}{*}{DINO~\cite{zhang2022dino}} 
         & R-50~\cite{resnet} & 47.7M & 244G & 49.2 & 66.7 & 53.8 \\
         & Swin-T~\cite{swin} & 48.2M & 252G & 51.3 & 69.0 & 55.9 \\         
         & Swin-S~\cite{swin} & 69.5M & 332G & 53.0 & 71.2 & 57.6\\
         \rowcolor{mygray}
          \cellcolor{white} & SMT-S & \textbf{39.9M} & 309G & \textbf{54.0} & \textbf{71.9} & \textbf{59.0}   \\

\bottomrule
\end{tabular}
\caption{A comparison of models with three different object detection frameworks.}
\label{tab:C2}

\end{table}

\subsection{Image Classification on ImageNet-1K}
\paragraph{Setup}
We conduct an evaluation of our proposed model and compare it with various networks on ImageNet-1K classification~\cite{imagenet1k}. To ensure a fair comparison, we follow the same training recipes as previous works~\cite{deit, swin,shunted}. Specifically, we train the models for 300 epochs with an image size of $224\times224$ and report the top-1 validation accuracy. The batch size used is 1024, and we employ the AdamW optimizer~\cite{adam, adamw} with a weight decay of 0.05 and a learning rate of $1\times10^{-3}$. In addition, we investigate the effectiveness of SMTs when pretrained on ImageNet-22K.(Further details regarding the training process can be found in Appendix~\ref{Appendix:B})

\paragraph{Results}
Tab.~\ref{exp:imagenet} presents a comparison of our proposed SMT with various models, and the results demonstrate that our models outperform various architectures with fewer parameters and lower computation costs. 
Specifically, concerning the tiny-sized model, SMT achieves an impressive top-1 accuracy of 82.2\%, surpassing PVTv2-b1~\cite{pvtv2} and Shunted-T~\cite{shunted} by significant margins of 3.5\% and 2.4\%, respectively.
Furthermore, when compared to small-sized and base-sized models, SMT maintains its leading position. Notably, SMT-B achieves a top-1 accuracy of 84.3\% with only 32M parameters and 7.7GFLOPs of computation, outperforming many larger models such as Swin-B~\cite{swin}, ConvNeXt-B~\cite{convnext}, and FocalNet-B~\cite{focalnet}, which have over 70M parameters and 15GFLOPs of computation. Additionally, to evaluate the scalability of the SMT, we have also created smaller and larger models, and the experimental results are presented in the Appendix~\ref{Appendix:C}.

We also report the ImageNet-22K pre-training results here in Tab.~\ref{imagenet22k}.
When compared to the previously best results, our models achieve significantly better accuracy with a reduced number of parameters and FLOPs.
SMT-L attains an 88.1\% top-1 accuracy, surpassing InternImage-XL by 0.1\% while utilizing significantly fewer parameters (80.5M vs. 335M) and exhibiting lower FLOPs (54.6G vs. 163G). 
This highly encouraging outcome underscores the impressive scalability capabilities of SMT.

\subsection{Object Detection and Instance Segmentation}
\paragraph{Setup}
We make comparisons on object detection with COCO 2017~\cite{coco}. We use SMT-S/B pretrained on ImageNet-1K as the foundation for three well-known object detectors: Mask R-CNN~\cite{maskrcnn}, Cascade Mask R-CNN~\cite{cai2018cascade}, and RetinaNet~\cite{lin2017focal}. To demonstrate a consistent comparison,  two training schedules ($1\times$ schedule with 12 epochs and $3\times$ schedule with 36 epochs) are adopted in Mask R-CNN. In $3\times$ schedule, we use a multi-scale training strategy by randomly resizing the shorter side of an image to between [480, 800]. 
We take AdamW optimizer with a weight decay of 0.05 and an initial learning rate of $2\times10^{-4}$. Both models are trained with batch size 16. To further showcase the versatility of SMT, we conducted a performance evaluation of SMT with three other prominent object detection frameworks, namely Sparse RCNN~\cite{sun2021sparse}, ATSS~\cite{zhang2020bridging}, and DINO~\cite{zhang2022dino}. We initialize the backbone with weights pre-trained on ImageNet-1K and fine-tune the model using a 3$\times$ schedule for Sparse RCNN and ATSS.

\paragraph{Results}
Tab.~\ref{tab:maskrcnn} presents the superior performance of SMT over other networks with Mask R-CNN~\cite{maskrcnn} under various model sizes. Specifically, SMT demonstrates a significant improvement in box mAP of 5.6 and 4.2 over Swin Transformer in 1$\times$ schedule under small and base model sizes, respectively. Notably, with 3$\times$ schedule and multi-scale training, SMT still consistently outperforms various backbones. For instance segmentation, the results also demonstrate that our SMT achieves higher mask mAP in comparison to previous SOTA networks. In particular, for small and base models in the 1$\times$ schedule, we achieve 1.5 and 0.9 points higher than FocalNet, respectively. Furthermore, to assess the generality of SMT, we trained two additional detection models, Cascade Mask R-CNN~\cite{cai2018cascade} and RetinaNet~\cite{lin2017focal}, using SMT-S as the backbone. The results, presented in Tab.~\ref{tab:cascade}, show clear improvements over various backbones in both box and mask mAPs. The resulting box mAPs for Sparse R-CNN, ATSS and DINO are presented in Tab.~\ref{tab:C2}, which indicate that SMT outperforms other networks consistently across all detection frameworks, highlighting its exceptional performance in downstream tasks.

\subsection{Semantic Segmentation on ADE20K}
\paragraph{Setup}
We evaluate the SMT for semantic segmentation using the ADE20K dataset. To conduct the evaluation, we use UperNet as the segmentation method and closely followed the training settings proposed by ~\cite{swin}. Specifically, we train UperNet~\cite{upernet} for 160k iterations with an input resolution of $512\times512$. We employ the AdamW optimizer with a weight decay of 0.01, and set the learning rate to $6\times10^{-5}$.

\paragraph{Results}
The results are presented in Tab.~\ref{tab:seg}, which shows that our SMT outperforms Swin, FocalNet, and Shunted Transformer significantly under all settings. Specifically, SMT-B achieves 1.5 and 0.9 mIoU gains compared to Swin-B and a 0.6 and 0.1 mIoU improvement over Focal-B at single- and multi-scale, respectively, while consuming significantly fewer FLOPs and reducing the model size by more than 50\%. Even for the SMT with a small model size, it achieves comparable accuracy with the previous SOTA models which have a larger model size.

\begin{table}[h]
\setlength{\tabcolsep}{2pt}
\centering
\resizebox{0.95\linewidth}{!}{
  \begin{tabular}{l|cccc}
    \toprule
    Backbone & \#Param(M) & FLOPs(G) & $mIoU_{ss}$ & $mIoU_{ms}$ \\
    \midrule	 
    ResNet-101~\cite{resnet} & 86 & 1029 & 44.9 & - \\
    DeiT-S~\cite{deit} & 52 & 1099 & 44.0 & - \\
    Swin-T~\cite{swin} & 60 & 941 & 44.5 & 45.8  \\
    Focal-T~\cite{focal}& 62 & 998 & 45.8 & 47.0 \\
    FocalNet-T~\cite{focal} & 61 & 949 & 46.8 & 47.8 \\    
    Swin-S~\cite{swin}  & 81 & 1038 & 47.6 & 49.5  \\
    ConvNeXt-S~\cite{convnext} & 82 & 1027 & 49.6 & - \\
    Shunted-S~\cite{shunted}  & 52 & 940 & 48.9 & 49.9  \\
    FocalNet-S~\cite{focalnet}   & 84 & 1044 & 49.1 & 50.1 \\    
    Focal-S~\cite{focal}  & 85 & 1130 & 48.0 & 50.0\\ 
    Swin-B~\cite{swin} & 121 & 1188 & 48.1 & 49.7  \\   
    Twins-SVT-L~\cite{twins} & 133 & - & 48.8 & 50.2 \\
    Focal-B~\cite{focal}  & 126 & 1354 & 49.0 & 50.5 \\  
    \midrule
    \textbf{SMT-S} & 50.1 & 935 & 49.2 & 50.2 \\ 
    \textbf{SMT-B} & 61.8 & 1004 & \textbf{49.6} & \textbf{50.6} \\  
    \bottomrule

  \end{tabular} 
  }
  \caption{Semantic segmentation on ADE20K~\cite{ade20k}. All models are trained with UperNet~\cite{upernet}. $mIoU_{ms}$ means multi-scale evaluation.}
  \label{tab:seg}\vspace{-4mm}
\end{table}

\subsection{Ablation Study}

\paragraph{Number of heads in Multi-Head Mixed Convolution}
Table~\ref{ablation:heads} shows the impact of the number of convolution heads in the Multi-Head Mixed Convolution (MHMC) on our model's performance. The experimental results indicate that while increasing the number of diverse convolutional kernels is advantageous for modeling multi-scale features and expanding the receptive field, adding more heads introduces larger convolutions that may negatively affect network inference speed and reduce throughput. Notably, we observed that the top-1 accuracy on ImageNet-1K peaks when the number of heads is 4, and increasing the number of heads does not improve the model's performance. This findings suggest that introducing excessive distinct convolutions or using a single convolution is not suitable for our SMT, emphasizing the importance of choosing the appropriate number of convolution heads to model a specific degree of multi-scale spatial features.
\begin{table}[h]
\setlength\tabcolsep{0.8pt}
    \small
    \centering
    \begin{tabular}{c|cccc}
    \toprule
    Heads Number & Params(M) & FLOPs(G) & top-1 (\%) & \makecell{throughput\\(images/s)} \\
    \hline
    1 & 11.5 & 2.4 & 81.8 & 983  \\
    2 & 11.5 & 2.4 & 82.0 & 923  \\
    4 & 11.5 & \textbf{2.4} & \textbf{82.2} & 833 \\
    6 & 11.6 & 2.5 & 81.9 & 766  \\
    8 & 11.6 & 2.5 & 82.0 & 702 \\

    \bottomrule
    \end{tabular}
    \caption{Model performance with number of heads in MHMC. We analyzed the model's performance for the number of heads ranging from 1 to 8. Throughput is measured using a V100 GPU, following~\cite{swin}.}
    \label{ablation:heads}\vspace{-6mm}
\end{table}

\paragraph{Different aggregation strategies}
After applying the MHMC, we introduce an aggregation module to achieve information fusion. Table~\ref{tab:aggregation} presents a comparison of different aggregation strategies, including a single linear layer, two linear layers, and an Invert BottleNeck (IBN)~\cite{mobilenetv2}. Our proposed scale-aware aggregation (SAA) consistently outperforms the other fusion modules, demonstrating the effectiveness of SAA in modeling multi-scale features with fewer parameters and lower computational costs. Notably, as the size of the model increases, our SAA can exhibit more substantial benefits while utilizing a small number of parameters and low computational resources.
\begin{table}[h]
\setlength\tabcolsep{5pt}
    \small
    \centering
    \begin{tabular}{c|ccc}
    \toprule
    Aggregation Strategy & \makecell{Params\\(M)} & \makecell{FLOPs\\(G)} & \makecell{top-1\\(\%)} \\
    \hline
    No aggregation & 10.9 & 2.2 & 81.5  \\
    \hline
    Single Linear ($c \rightarrow c$) & 11.2 & 2.3 & 81.6  \\
    Two Linears ($c \rightarrow c \rightarrow c$) & 11.5 & 2.4 & 81.9  \\
    IBN ($c \rightarrow 2c \rightarrow c$) & 12.1 & 2.6 & 82.1  \\
    SAA($c \rightarrow 2c \rightarrow c$) & 11.5 & 2.4 & \textbf{82.2} \\

    \bottomrule
    \end{tabular}
    \caption{Model performance for different aggregation methods.}
    \label{tab:aggregation}\vspace{-6mm}
\end{table}

\paragraph{Different hybrid stacking strategies}
\label{exp:ehs}
In Sec.~\ref{EHS}, we propose two hybrid stacking strategies to enhance the modeling of the transition from local to global dependencies. 
The results shown in Table~\ref{tab:ehs} indicate that the first strategy which sequentially stacks one scale-aware modulation block and one multi-head self-attention block is better, achieving a performance gain of 0.3\% compared to the other strategy. Furthermore, the strategy stacking all MSA blocks achieves comparable performance as well, which means retaining the MSA block in the last two stages is crucial.

\begin{table}[h]
\renewcommand{\arraystretch}{1.2}
\setlength\tabcolsep{1.5pt}
    \small
    \centering
    \begin{tabular}{c|c|ccc}
    \toprule
    Stacking Strategy & Hybrid & \makecell{Params\\(M)} & \makecell{FLOPs\\(G)} & \makecell{top-1\\(\%)} \\
    \hline
    $(SAM \times N)$ & \XSolidBrush & 11.8 & 2.5 & 81.4  \\
    $(MSA \times N)$ & \XSolidBrush & 11.2 & 2.3 & 81.8  \\
    \hline
    $(SAM \times 1 + MSA \times 1) \times \frac{N}{2}$ & \Checkmark & 11.5 & 2.4 & \textbf{82.2}  \\
    $(SAM \times \frac{N}{2} + MSA \times \frac{N}{2})$ & \Checkmark & 11.5 & 2.4 & 81.9  \\

    \bottomrule
    \end{tabular}
    \vspace{2mm}
    \caption{Top-1 accuracy on ImageNet-1K of different stacking strategies.}
    \label{tab:ehs}\vspace{-4mm}
\end{table}

\paragraph{Component Analysis}
In this section, we investigate the individual contributions of each component by conducting an ablation study on SMT. Initially, we employ a single-head convolution module and no aggregation module to construct the modulation. Based on this, we build an attention-free network, which can achieve 80\% top-1 accuracy on the ImageNet-1K dataset. 
The effects of all the proposed methods on the model's performance are given in Tab.~\ref{as:component_analysis}, which can be summarized as followings.

\begin{itemize}[leftmargin=*]
    \item \textbf{Multi-Head Mixed Convolution (MHMC) }To enhance the model's ability to capture multi-scale spatial features and expand its receptive field, we replaced the single-head convolution with our proposed MHMC. This module proves to be effective for modulation, resulting in a 0.8\% gain in accuracy.
    \item \textbf{Scale-Aware Aggregation (SAA) }We replace the single linear layer with our proposed scale-aware aggregation. The SAA enables effective aggregation of the multi-scale features captured by MHMC. Building on the previous modification, the replacement leads to a 1.6\% increase in performance.
    \item \textbf{Evolutionary Hybrid Network (EHN)}
    We incorporate the self-attention module in the last two stages of our model, while also implementing our proposed hybrid stacking strategy in the penultimate stage, which improves the modeling of the transition from local to global dependencies as the network becomes deeper, resulting in a significant gain of 2.2\% in performance based on the aforementioned modifications.
\end{itemize}

\begin{table}
\setlength\tabcolsep{3pt}
    \small
    \centering
    \begin{tabular}{ccc|ccc}
        \toprule
        \makecell{MHMC}& \makecell{ SAA }  & \makecell{ EHN } & Params(M) & FLOPs(G) & top-1 (\%) \\
        \hline
         & &  &11.1  &2.3 & 80.0 ({$\uparrow$0.0})\\
        \checkmark & &  &11.2  &2.3 & 80.8 ({$\uparrow$0.8}) \\
        \checkmark &\checkmark &  &12.1  &2.5 & 81.6 ({$\uparrow$1.6}) \\
        \checkmark &\checkmark &\checkmark &11.5  &2.4 & 82.2 ({$\uparrow$2.2}) \\
        \bottomrule
    \end{tabular}
    \vspace{2mm}
    \caption{Component analysis for SMT. Three variations are gradually added to the original attention-free network.}
    \label{as:component_analysis}\vspace{-4mm}
\end{table}

\section{Conclusion}
In this paper, we introduce a new hybrid ConvNet and vision Transformer backbone, namely Scale-Aware Modulation Transformer (SMT), which can effectively simulate the transition from local to global dependencies as the network becomes deeper, resulting in superior performance. To satisfy the requirement of foundation models, we propose a new Scale-Aware Modulation that includes a potent multi-head mixed convolution module and a lightweight scale-aware aggregation module. Extensive experiments demonstrate the efficacy of SMT as a backbone for various downstream tasks, achieving comparable or better performance than well-designed ConvNets and vision Transformers, with fewer parameters and FLOPs. We anticipate that the exceptional performance of SMT on diverse vision problems will encourage its adoption as a promising new generic backbone for efficient visual modeling.
\vspace{2mm}
\section*{Acknowledgement}
This research is supported in part by NSFC (Grant No.: 61936003), Alibaba DAMO Innovative Research Foundation  (20210925), Zhuhai Industry Core, Key Technology Research Project (no. 2220004002350) and National Key Research and Development Program of China (2022YFC3301703). We thank the support from the Alibaba-South China University of Technology Joint Graduate Education Program.

\clearpage
{\small
\bibliographystyle{ieee_fullname}
\bibliography{egbib}
}

\clearpage
\noindent\textbf{\Large Appendix}
\appendix
\section{Detailed Architecture Specifications} \label{Appendix:A}
Tab.~\ref{tab:arch-spec} provides a detailed overview of the architecture specifications for all models, with an assumed input image size of $224\times224$. The stem of the model is denoted as "conv $n \times n$, 64-d, BN; conv $2\times2$, 64-d, LN", representing two convolution layers with a stride of 2 to obtain a more informative token sequence with a length of $\frac{H}{4} \times \frac{W}{4}$. Here, "BN" and "LN" indicate Batch Normalization and Layer Normalization~\cite{ba2016layer}, respectively, while "64-d" denotes the convolution layer with an output dimension of 64.
The multi-head mixed convolution module with 4 heads (conv $3\times3$, conv $5\times5$, conv $7\times7$, conv $9\times9$) is denoted as "sam. head. 4", while "msa. head. 8" represents the multi-head self-attention module with 8 heads. Additionally, "sam. ep\_r. 2" indicates a Scale-Aware Aggregation module with twice as much expanding ratio.

\vspace{-2mm}
\section{Detailed Experimental Settings} \label{Appendix:B}
\subsection{Image classification on ImageNet-1K}
We trained all models on the ImageNet-1K dataset~\cite{imagenet1k} for 300 epochs, using an image size of $224 \times 224$ . Following Swin~\cite{swin}, we utilized a standardized set of data augmentations~\cite{cubuk2020randaugment}, including Random Augmentation, Mixup~\cite{zhang2017mixup}, CutMix~\cite{yun2019cutmix}, and Random Erasing~\cite{zhong2020random}. To regularize our models, we applied Label Smoothing~\cite{szegedy2016rethinking} and DropPath~\cite{huang2016deep} techniques. The initial learning rate for all models was set to $2 \times 10^{-3}$ after 5 warm-up epochs, beginning with a rate of $1 \times 10^{-6}$. To optimize our models, we employed the AdamW~\cite{adamw} algorithm and a cosine learning rate scheduler~\cite{loshchilov2016sgdr}. The weight decay was set to 0.05 and the gradient clipping norm to 5.0. For our mini, tiny, small, base, and large models, we used stochastic depth drop rates of 0.1, 0.1, 0.2, 0.3, and 0.5, respectively. For more details, please refer to the Tab.~\ref{tab:B1} provided.

\begin{table}[h]
\footnotesize
\centering
\begin{tabular}{l|l}
\toprule
config & \vspace{0.5mm} value \\
\midrule
optimizer & \vspace{0.5mm} AdamW \\
LR & \vspace{0.5mm}  2e-3 \\
weight decay & \vspace{0.5mm}  0.05 \\
optimizer momentum & \vspace{0.5mm}  $\beta_1, \beta_2{=}0.9, 0.999$ \\
batch size & \vspace{0.5mm}  1024 \\
LR schedule & \vspace{0.5mm}  cosine \\
minimum learning rate & \vspace{0.5mm}  1e-5 \\
warmup epochs & \vspace{0.5mm}  5 \\
warmup learning rate & \vspace{0.5mm}  1e-6 \\
training epochs & \vspace{0.5mm}  300  \\
augmentation & \vspace{0.5mm}  rand-m9-mstd0.5-inc1 \\
color jitter & \vspace{0.5mm}  0.4 \\
mixup $\alpha$ & \vspace{0.5mm}  0.2 \\
cutmix $\alpha$ & \vspace{0.5mm}  1.0 \\
random erasing & \vspace{0.5mm}  0.25 \\ 
label smoothing & \vspace{0.5mm}  0.1 \\
gradient clip & \vspace{0.5mm}  5.0 \\
drop path & \vspace{0.5mm}  [0.1, 0,1, 0,2, 0,3, 0.5] (M,T,S,B,L) \\
\bottomrule
\end{tabular}
\caption{Image Classification Training Settings}
\label{tab:B1}
\end{table}

\subsection{Image classification pretrained on ImageNet-22K}
We trained the SMT-L model for 90 epochs using a batch size of 4096 and an input resolution of 224×224. The initial learning rate was set to $1\times10^{-3}$ after a warm-up period of 5 epochs. The stochastic depth drop rates were set to 0.1. Following pretraining, we performed fine-tuning on the ImageNet-1K dataset for 30 epochs. The initial learning rate was set to $2\times10^{-5}$, and we utilized a cosine learning rate scheduler and AdamW optimizer. The stochastic depth drop rate remained at 0.1 during fine-tuning, while both CutMix and Mixup augmentation techniques were disabled.

\vspace{-2mm}
\subsection{Object Detection and Instance Segmentation}
In transferring SMT to object detection and instance segmentation on COCO~\cite{coco}, we have considered six common frameworks: Mask R-CNN~\cite{maskrcnn}, Cascade Mask RCNN~\cite{cai2018cascade}, RetinaNet~\cite{lin2017focal}, Sparse R-CNN~\cite{sun2021sparse}, ATSS~\cite{zhang2020bridging}, and DINO~\cite{zhang2022dino}. For DINO, the model is fine-tuned for 12 epochs, utilizing 4 scale features. For optimization, we adopt the AdamW optimizer with an initial learning rate of 0.0002 and a batch size of 16. When training models of different sizes, we adjust the training settings according to the settings used in image classification. The detailed hyper-parameters used in training models are presented in Tab.~\ref{tab:B2}.

\begin{table}[h]
\setlength\tabcolsep{0.8pt}
\footnotesize
\centering
\begin{tabular}{l|l}
\toprule
config & \vspace{0.5mm} value \\
\midrule
optimizer & \vspace{0.5mm} AdamW \\
LR & \vspace{0.5mm}  0.0002 \\
weight decay & \vspace{0.5mm}  0.05 \\
optimizer momentum & \vspace{0.5mm}  $\beta_1, \beta_2{=}0.9, 0.999$ \\
batch size & \vspace{0.5mm}  16 \\
LR schedule & \vspace{0.5mm} steps:[8, 11] ($1\times$), [27, 33] ($3\times$) \\
warmup iterations (ratio) & \vspace{0.5mm} 500 (0.001) \\
training epochs & \vspace{0.5mm} 12 ($1\times$), 36 ($3\times$)  \\
scales & \vspace{0.5mm} (800, 1333) ($1\times$), Multi-scales~\cite{swin} ($3\times$) \\
drop path & \vspace{0.5mm} 0.2 (Small), 0.3 (Base) \\
\bottomrule
\end{tabular}
\caption{Object Detection and Instance Segmentation Training Settings}
\label{tab:B2}
\vspace{-1mm}
\end{table}

\begin{table*}[t]\footnotesize
\setlength\tabcolsep{2pt}
\centering
\begin{tabular}{c|c|c|c|c|c|c|c}
\toprule
 & \begin{tabular}[c]{@{}c@{}}downsp. rate \\ (output size)\end{tabular} & Layer Name &SAM-M  & SAM-T & SAM-S & SAM-B & SAM-L \\
\midrule
\multirow{4}{*}{stage 1} & 
\multirow{4}{*}{\begin{tabular}[c]{@{}c@{}}4$\times$\\ (56$\times$56)\end{tabular}} & 
\multirow{4}{*}{\makecell[c]{SAM \\ Block}} & 
\makecell[c]{conv 3$\times$3, 64-d, BN \\ conv 2$\times$2, 64-d, LN}  & 
\makecell[c]{conv 3$\times$3, 64-d, BN \\ conv 2$\times$2, 64-d, LN}  & 
\makecell[c]{conv 7$\times$7, 64-d, BN \\ conv 2$\times$2, 64-d, LN}  & 
\makecell[c]{conv 7$\times$7, 64-d, BN \\ conv 2$\times$2, 64-d, LN}  & 
\makecell[c]{conv 7$\times$7, 96-d, BN \\ conv 2$\times$2, 96-d, LN}  \\
\cline{4-8}
& & & $\begin{bmatrix}\text{dim 64}\\\text{sam.head. 4}\\\text{sam.ep\_r. 2}\end{bmatrix}$ $\times$ 1   & 
$\begin{bmatrix}\text{dim 64}\\\text{sam.head. 4}\\\text{sam.ep\_r. 2}\end{bmatrix}$ $\times$ 2    & 
$\begin{bmatrix}\text{dim 64}\\\text{sam.head. 4}\\\text{sam.ep\_r. 2}\end{bmatrix}$ $\times$ 3   & 
$\begin{bmatrix}\text{dim 64}\\\text{sam.head. 4}\\\text{sam.ep\_r. 2}\end{bmatrix}$ $\times$ 4   & 
$\begin{bmatrix}\text{dim 96}\\\text{sam.head. 4}\\\text{sam.ep\_r. 2}\end{bmatrix}$ $\times$ 4   \\
\hline
\multirow{4}{*}{stage 2}  & 
\multirow{4}{*}{\begin{tabular}[c]{@{}c@{}}8$\times$\\ (28$\times$28)\end{tabular}} & 
\multirow{4}{*}{\makecell[c]{SAM \\ Block}} & 
conv 3$\times$3, 128-d, LN & 
conv 3$\times$3, 128-d, LN & 
conv 3$\times$3, 128-d, LN & 
conv 3$\times$3, 128-d, LN &
conv 3$\times$3, 192-d, LN \\
\cline{4-8}
& & & $\begin{bmatrix}\text{dim 128}\\\text{sam.head. 4}\\\text{sam.ep\_r. 2}\end{bmatrix}$ $\times$ 1  & 
$\begin{bmatrix}\text{dim 128}\\\text{sam.head. 4}\\\text{sam.ep\_r. 2}\end{bmatrix}$ $\times$ 2 & 
$\begin{bmatrix}\text{dim 128}\\\text{sam.head. 4}\\\text{sam.ep\_r. 2}\end{bmatrix}$ $\times$ 4 & 
$\begin{bmatrix}\text{dim 128}\\\text{sam.head. 4}\\\text{sam.ep\_r. 2}\end{bmatrix}$ $\times$ 6 &
$\begin{bmatrix}\text{dim 192}\\\text{sam.head. 4}\\\text{sam.ep\_r. 2}\end{bmatrix}$ $\times$ 6 \\
\hline
\multirow{5}{*}{stage 3}  & 
\multirow{5}{*}{\begin{tabular}[c]{@{}c@{}}16$\times$\\ (14$\times$14)\end{tabular}}  & 
\multirow{5}{*}{\makecell[c]{Mix \\ Block}} & 
conv 3$\times$3, 256-d , LN & 
conv 3$\times$3, 256-d , LN & 
conv 3$\times$3, 256-d , LN & 
conv 3$\times$3, 256-d , LN &
conv 3$\times$3, 384-d , LN \\
\cline{4-8}
& & & $\begin{bmatrix}\text{dim 256}\\\text{sam.head. 4}\\\text{sam.ep\_r. 2}\\\text{msa.head. 8}\end{bmatrix}$ $\times$ 4 & 
$\begin{bmatrix}\text{dim 256}\\\text{sam.head. 4}\\\text{sam.ep\_r. 2}\\\text{msa.head. 8}\end{bmatrix}$ $\times$ 8 & 
$\begin{bmatrix}\text{dim 256}\\\text{sam.head. 4}\\\text{sam.ep\_r. 2}\\\text{msa.head. 8}\end{bmatrix}$ $\times$ 18  & 
$\begin{bmatrix}\text{dim 256}\\\text{sam.head. 4}\\\text{sam.ep\_r. 2}\\\text{msa.head. 8}\end{bmatrix}$ $\times$ 28 &
$\begin{bmatrix}\text{dim 384}\\\text{sam.head. 4}\\\text{sam.ep\_r. 2}\\\text{msa.head. 8}\end{bmatrix}$ $\times$ 28 \\
\hline
\multirow{3}{*}{stage 4} & 
\multirow{3}{*}{\begin{tabular}[c]{@{}c@{}}32$\times$\\ (7$\times$7)\end{tabular}}  & 
\multirow{3}{*}{\makecell[c]{MSA \\ Block}} & 
conv 3$\times$3, 512-d , LN & 
conv 3$\times$3, 512-d , LN & 
conv 3$\times$3, 512-d , LN  & 
conv 3$\times$3, 512-d , LN  & 
conv 3$\times$3, 768-d , LN \\
\cline{4-8}
& & & $\begin{bmatrix}\text{dim 512}\\\text{msa.head 16}\end{bmatrix}$ $\times$ 1 & 
$\begin{bmatrix}\text{dim 512}\\\text{msa.head 16}\end{bmatrix}$ $\times$ 1  & 
$\begin{bmatrix}\text{dim 512}\\\text{msa.head 16}\end{bmatrix}$ $\times$ 1 & 
$\begin{bmatrix}\text{dim 512}\\\text{msa.head 16}\end{bmatrix}$ $\times$ 2 &
$\begin{bmatrix}\text{dim 768}\\\text{msa.head 16}\end{bmatrix}$ $\times$ 3 \\
\bottomrule
\end{tabular}
\normalsize
\caption{Detailed architecture specifications at four stages for SMT.}
\label{tab:arch-spec}
\end{table*}

\begin{figure*}[t]
    \centering
        \includegraphics[width=0.85\textwidth]{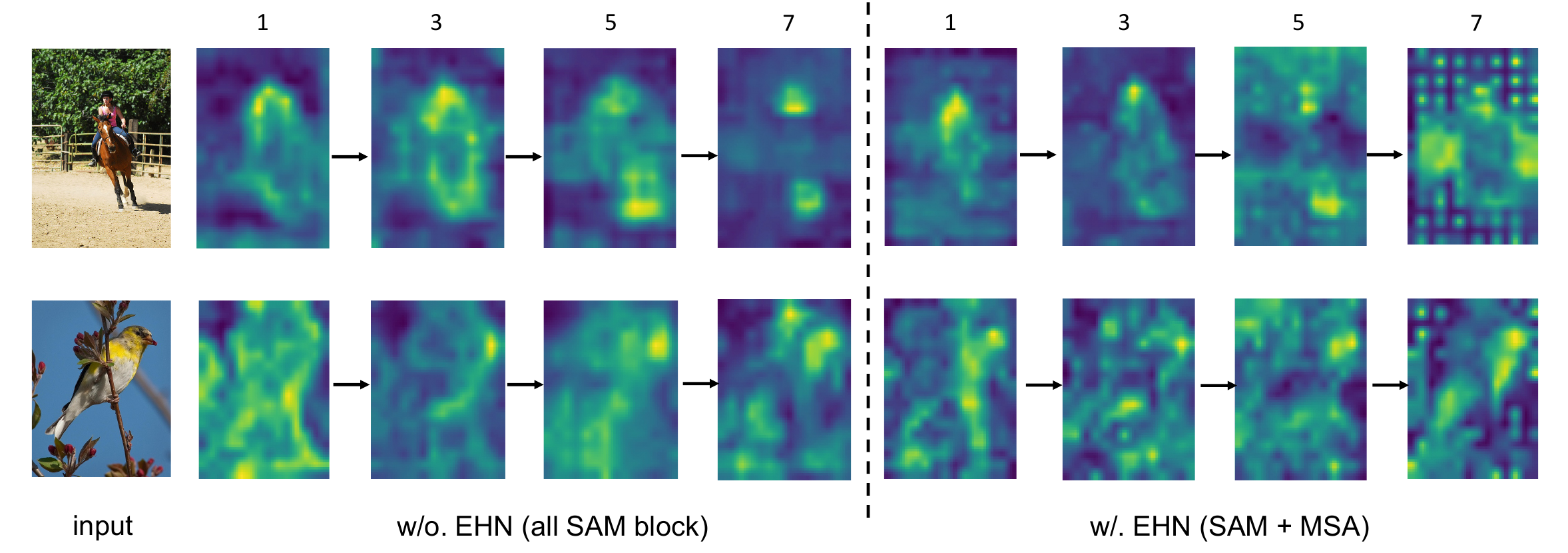}
    \caption{Visualization of modulation values at the penultimate stage for two variants of SMT. \textbf{(Left: w/o. EHN)} 
 Stacking of SAM blocks exclusively in the penultimate stage. \textbf{(Right: w/. EHN)} The utilization of an evolutionary hybrid stacking strategy, wherein one SAM block and one MSA are successively stacked.}
  \label{fig:sup_ehn}
  \vspace{-2mm}
\end{figure*}

\vspace{-4mm}
\subsection{Semantic Segmentation}
For ADE20K, we utilized the AdamW optimizer with an initial learning rate of 0.00006, a weight decay of 0.01, and a batch size of 16 for all models trained for 160K iterations. In terms of testing, we reported the results using both single-scale (SS) and multi-scale (MS) testing in the main comparisons. For multi-scale testing, we experimented with resolutions ranging from 0.5 to 1.75 times that of the training resolution. To set the path drop rates in different models, we used the same hyper-parameters as those used for object detection and instance segmentation.

\vspace{-2mm}
\section{More Experiments}
\label{Appendix:C}
\subsection{More Variants of SMT}
This section demonstrates how we scaled our SMT to create both smaller (SMT-M) and larger (SMT-L) models. Their detailed architecture settings are provided in Tab.~\ref{tab:arch-spec}, along with previous variants. We then evaluated their performance on the ImageNet-1K dataset.

As shown in Tab.~\ref{tab:C1}, SMT-M achieves competitive results with a top-1 accuracy of 78.4\%, despite having only 6.5M parameters and 1.3 GFLOPs of computation. On the other side, SMT-L shows an example to scale our SMT to larger models, which outperforms other state-of-the-art networks with similar parameters and computation costs, achieving a top-1 accuracy of 84.6\%. These results confirm the strong scalability of the SMT architecture, which can be applied to create models of varying sizes, demonstrating its immense potential.

\begin{table}[h]
\setlength\tabcolsep{2.5pt}
\centering
\small
\begin{tabular}{c|ccc|c}
\Xhline{1.0pt}
method & \begin{tabular}[c]{@{}c@{}}image \\ size\end{tabular} & \#param. & FLOPs & \begin{tabular}[c]{@{}c@{}}ImageNet \\ top-1 acc.\end{tabular} \\
\hline
RegNetY-4G~\cite{regnet} & 224$^2$ & 21M & 4.0G & 80.0 \\
RegNetY-8G~\cite{regnet} & 224$^2$ & 39M & 8.0G & 81.7 \\
RegNetY-16G~\cite{regnet} & 224$^2$ & 84M & 16.0G & 82.9 \\
EffNet-B3 ~\cite{efficientnet} & 300$^2$ & 12M & 1.8G &  81.6 \\
EffNet-B4 ~\cite{efficientnet} & 380$^2$ & 39M & 4.2G &  82.9 \\
EffNet-B5 ~\cite{efficientnet} & 456$^2$ & 30M & 9.9G &  83.6 \\
EffNet-B6 ~\cite{efficientnet} & 528$^2$ & 43M & 19.0G &  84.0 \\
PVT-T~\cite{pvt} & 224$^2$ & 13M & 1.8G & 75.1 \\
PVT-S~\cite{pvt} & 224$^2$ & 25M & 3.8G & 79.8 \\
PVT-M~\cite{pvt} & 224$^2$ & 44M & 6.7G & 81.2 \\
PVT-L~\cite{pvt} & 224$^2$ & 61M & 9.8G & 81.7 \\
Swin-T~\cite{swin} & 224$^2$ & 29M & 4.5G & 81.3 \\
Swin-S~\cite{swin} & 224$^2$ & 49.6M & 8.7G & 83.0 \\
Swin-B~\cite{swin} & 224$^2$ & 87.8M & 15.4G &  83.4 \\
Twins-S~\cite{twins} & 224$^2$ & 24M & 2.9G &  81.7 \\
Twins-B~\cite{twins} & 224$^2$ & 56M & 8.6G &  83.2 \\
Focal-T~\cite{focal} & 224$^2$ & 29M & 4.9G & 82.2 \\
Focal-B~\cite{focal} & 224$^2$ & 89.8M & 16.4G & 83.8 \\
Shunted-T~\cite{shunted} & 224$^2$ & 11.5M & 2.1G & 79.8 \\
Shunted-S~\cite{shunted} & 224$^2$ & 22.4M & 4.9G & 82.9 \\
Shunted-B~\cite{shunted} & 224$^2$ & 39.6M & 8.1G & 84.0 \\
FocalNet-T~\cite{focalnet} & 224$^2$ & 28.6M & 4.5G & 82.3 \\
FocalNet-S~\cite{focalnet} & 224$^2$ & 50.3M & 8.7G & 83.5 \\
FocalNet-B~\cite{focalnet} & 224$^2$ & 88.7M & 15.4G & 83.9 \\
MaxViT-T~\cite{tu2022maxvit} & 224$^2$ & 31M & 5.6G & 83.6 \\
MaxViT-S~\cite{tu2022maxvit} & 224$^2$ & 69M & 11.7G & 84.5 \\
MaxViT-B~\cite{tu2022maxvit} & 224$^2$ & 120M & 23.4G & 84.9 \\
\Xhline{1.0pt}
SMT-M & 224$^2$ & 6.5M & 1.3G & \textbf{78.4} \\
SMT-T & 224$^2$ & 11.5M & 2.4G & \textbf{82.2} \\
SMT-S & 224$^2$ & 20.5M & 4.7G & \textbf{83.7} \\
SMT-B & 224$^2$ & 32.0M & 7.7G & \textbf{84.3} \\
SMT-L & 224$^2$ & 80.5M & 17.7G & \textbf{84.6} \\

\Xhline{1.0pt}
\end{tabular}
\normalsize
\caption{Comparison of different backbones on ImageNet-1K classification. }
\label{tab:C1}
\vspace{-4mm}
\end{table}

\section{Additional Network Analysis}
\label{Appendix:D}
In Fig.~\ref{fig:sup_ehn}, we present the learned scale-aware modulation (SAM) value maps in two variants of SMT-T: evolutionary SMT, which employs an evolutionary hybrid stacking strategy, and general SMT, which only employs SAM in the penultimate stage. In evolutionary SMT-T, comprising a total of 8 layers in the penultimate stage, we select the layers $([1,3,5,7])$ containing SAM block and compare them with the corresponding layers in general SMT. Through visualization, we can observe some noteworthy patterns. In general SMT, the model primarily concentrates on local details in the shallow layers and on semantic information in the deeper layers. However, in evolutionary SMT, the focus region does not significantly shift as the network depth increases. Furthermore, it captures local details more effectively than general SMT in the shallow layers, while preserving detailed and semantic information about the target object at deeper layers. These results indicate that our evolutionary hybrid stacking strategy facilitates SAM blocks in capturing multi-granularity features while allowing multi-head self-attention (MSA) blocks to concentrate on capturing global semantic information. Accordingly, each block within each layer is more aptly tailored to its computational characteristics, leading to enhanced performance in diverse visual tasks.

\section{Additional Visual Examples}
\label{Appendix:E}
\begin{figure}[h]
    \centering
        \includegraphics[width=0.43\textwidth]{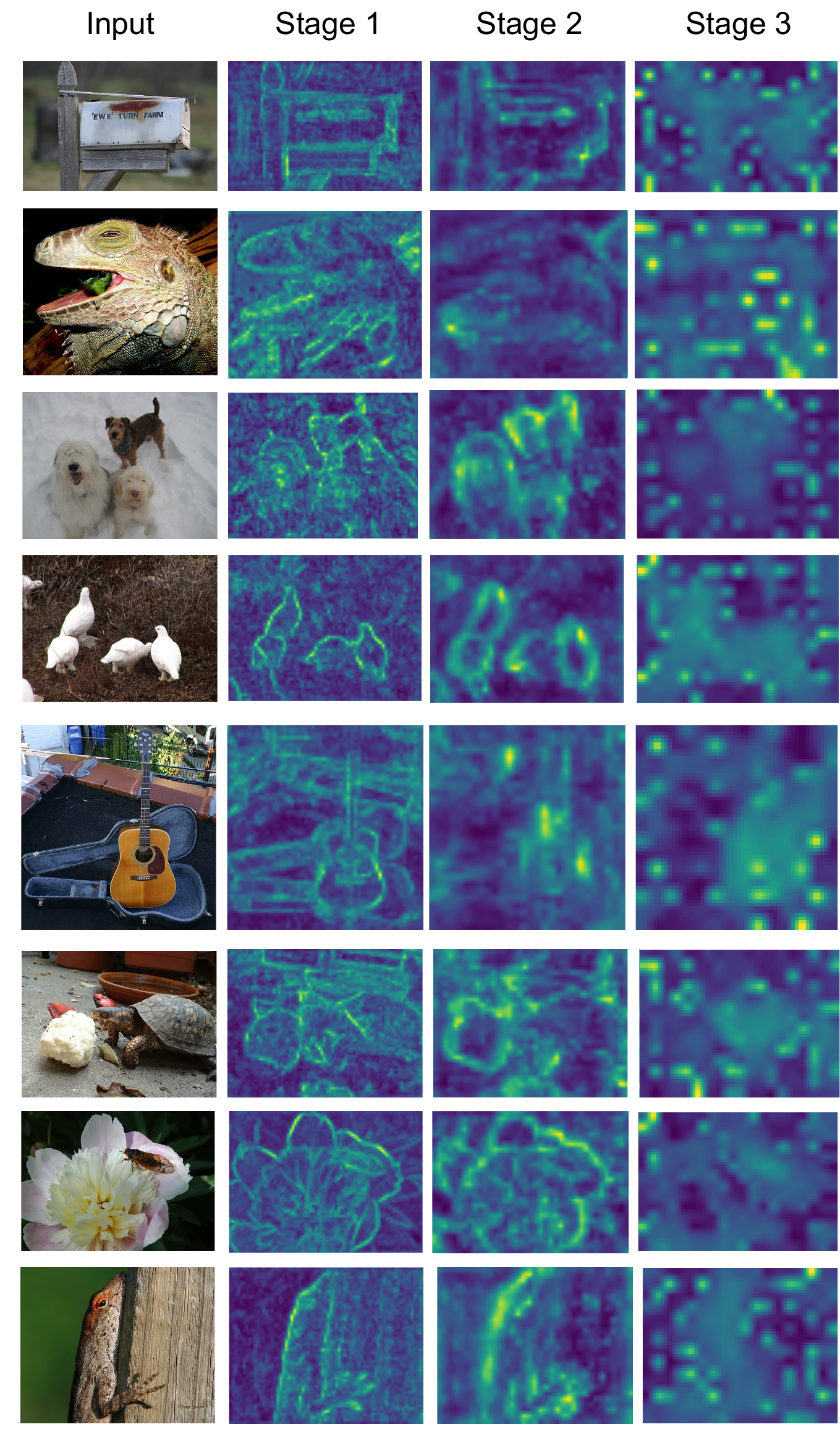}
    \caption{Visualization of modulation value maps at the top three stages.}
  \label{fig:adx}
\end{figure}

We present supplementary visualization of modulation value maps within our SMT. Specifically, we randomly select validation images from the ImageNet-1K dataset and generate visual maps for modulation at different stages, as illustrated in Fig~\ref{fig:adx}. The visualizations reveal that the scale-aware modulation is critical in strengthening semantically relevant low-frequency signals and accurately localizing the most discriminative regions within images. By exploiting this robust object localization capability, we can allocate more effort towards modulating these regions, resulting in more precise predictions. We firmly believe that both our multi-head mixed convolution module and scale-aware aggregation module have the potential to further enhance the modulation mechanism.

\end{document}